\pdfoutput=1

\documentclass[11pt]{article}
\usepackage[numbers]{natbib}

\usepackage{acl2025}

\usepackage{times}
\usepackage{latexsym}

\usepackage[T1]{fontenc}

\usepackage[utf8]{inputenc}

\usepackage{microtype}

\usepackage{inconsolata}

\title{Instructions for ACL 2023 Proceedings}

\usepackage{graphicx}
\usepackage{amsmath}
\usepackage{xspace}
\usepackage{subcaption}

\usepackage{tikz}
\usetikzlibrary{shapes, arrows.meta, positioning}

\usepackage{hyperref}       
\usepackage{url}            
\usepackage{booktabs}       
\usepackage{amsfonts}       
\usepackage{nicefrac}       
\usepackage{xcolor}         
\usepackage{enumerate}
\usepackage{cleveref}
\usepackage{tcolorbox}
\usepackage{wrapfig}

\newtcolorbox{mypromptbox}[2][]{colback=gray!5!white, colframe=gray!75!black, title=#2,#1, fontupper=\footnotesize}

\title{\promptevals: A Dataset of Assertions and Guardrails for Custom Production Large Language Model Pipelines}

\author{Reya Vir\thanks{Equal contribution.} \\
  UC Berkeley \\
  \texttt{reyavir@berkeley.edu} \\\And
  Shreya Shankar\footnotemark[1] \\
  UC Berkeley \\
  \texttt{shreyashankar@berkeley.edu} \\\And
  Harrison Chase \\
  LangChain \\
  \texttt{harrison@langchain.dev} \\\AND
  Will Fu-Hinthorn \\
  LangChain \\
  \texttt{will@langchain.dev} \\\And
  Aditya G. Parameswaran \\
  UC Berkeley \\
  \texttt{adityagp@berkeley.edu} \\
}

\begin{document}

\newcommand{\reya}[1]{\textcolor{red}{[Reya: #1]}}
\newcommand{\shreya}[1]{\textcolor{purple}{[Shreya: #1]}}
\newcommand{\agp}[1]{\textcolor{blue}{[Aditya: #1]}}
\newcommand{\promptevals}{\textsc{PromptEvals}\xspace}

\newcommand{\topic}[1]{\vspace{-3.5pt}\smallskip \smallskip \noindent{\bf #1.}}

\newcommand{\techreport}[1]{}
\newcommand{\papertext}[1]{#1}

\maketitle

\begin{abstract}
Large language models (LLMs) are increasingly deployed in specialized production data processing pipelines across diverse domains---such as finance, marketing, and e-commerce. However, when running them in production across many inputs, they often fail to follow instructions or meet developer expectations. To improve reliability in these applications, creating assertions or guardrails for LLM outputs to run alongside the pipelines is essential. Yet, determining the right set of assertions that capture developer requirements for a task is challenging. In this paper, we introduce \promptevals, a dataset of 2087 LLM pipeline prompts with 12623 corresponding assertion criteria, sourced from developers using our open-source LLM pipeline tools. This dataset is $5\times$ larger than previous collections. Using a hold-out test split of \promptevals as a benchmark, we evaluated closed- and open-source models in generating relevant assertions. Notably, our fine-tuned Mistral and Llama 3 models outperform GPT-4o by 20.93\% on average, offering both reduced latency and improved performance. We believe our dataset can spur further research in LLM reliability, alignment, and prompt engineering.
\end{abstract}

\section{Introduction}

Large language models (LLMs) have become increasingly popular for various data processing tasks. An open-source tool for building LLM pipelines, developed by some of the authors, now has over 3 million weekly downloads. Its community has created thousands of specialized prompts for diverse fields like medicine, finance, and sports, leveraging LLMs' impressive zero-shot and few-shot performance~\cite{brown2020language, Kojima2022LargeLM, Huang2022LanguageMA, Wei2021FinetunedLM}.

A common desire for developers using LLMs is to meet specific constraints on outputs, such as adhering to a particular structure or qualitative criteria~\cite{llmconstraints}. One approach to address this need is to collect large amounts of human preference data~\cite{ji2023beavertails, kopf2023openassistant, Zheng2023LMSYSChat1MAL}, and improve models through alignment techniques like supervised fine-tuning and reinforcement learning from human feedback~\cite{Chung2022ScalingIL, Ouyang2022TrainingLM, Wang2022SelfInstructAL}. However, these methods have a high barrier to entry, requiring dataset collection, model fine-tuning, and serving infrastructure, which is more complex than simply manipulating prompts for LLM calls. More importantly, fine-tuning isn't supervised at the constraint level---meaning that even fine-tuned LLMs often fail to consistently follow instructions that correspond to constraints in detailed prompts~\cite{kalai2023calibrated, Pan2023AutomaticallyCL, Huang2023ASO}.

An alternative solution involves implementing developer-specified {\em assertions} on LLM outputs~\cite{llmconstraints, shankar2024spade, rebedea2023nemo}. This approach typically follows two steps: first, defining binary evaluation criteria to represent the desired constraints; second, implementing these criteria as assertions to evaluate LLM outputs and resample these outputs when assertions fail.  However, developing effective assertion criteria is challenging---primarily due to the complexity of defining and conceptualizing these criteria, rather than their technical implementation~\cite{Kim2023EvalLMIE, shankar2024validates}. The complexity of coming up with assertions arises due to multiple factors: criteria can differ significantly between developers due to specific data, use cases, and end-user requirements~\cite{pmlr-v235-dong24c}; criteria must account for both user preferences and LLM-specific failure modes~\cite{shankar2024validates}; and developers may need to incorporate qualitative or subjective criteria that require LLMs themselves to perform the evaluation~\cite{evalullm, Kim2023EvalLMIE}. 

To improve custom and task-specific alignment for LLM pipelines, we need an approach that can examine developers' prompts and identify assertion criteria. These assertions can then be bolted onto the end of the LLM pipeline, allowing for automatic retrying of the pipeline if assertions fail. Developing such an approach requires substantial, diverse data on real-world LLM applications and their associated constraints. Fortunately, our open-source tool provides unique access to a diverse set of use-cases with associated prompts.

In this paper, we present \promptevals, a dataset created using our unique collection of real-world LLM prompts and use-cases. This dataset consists of 2087 human-contributed prompt templates for custom tasks and 12623 comprehensive assertion criteria. \promptevals has a median prompt template size of 191 tokens and is {\em more than five times larger than previous collections}~\cite{qin2024infobench, zhou2023instruction}. Our dataset and corresponding benchmark (20\% of the dataset) are hosted on HuggingFace\footnote{\url{https://huggingface.co/datasets/reyavir/PromptEvals}}. Using this benchmark, we evaluate GPT-4o and two open-source models, Llama 3-8b and Mistral-7b, on generating task-specific assertion criteria, and find that GPT-4o performs reasonably well out of the box, but its cost and latency to run for every prompt edit or pipeline update can be prohibitive in production environments---especially as prompts become increasingly complex and specialized. To address this, for our prompt engineering tools, we fine-tuned open-source models on our dataset (using Mistral-7b and Llama 3-8b architectures~\cite{jiang2023mistral, touvron2023llama}), and these models exceeded GPT-4o's F1 performance in identifying desirable assertions by 20.93\% on average. These fine-tuned models are made available to the community\footnote{\url{https://huggingface.co/reyavir/promptevals_mistral} and \url{https://huggingface.co/reyavir/promptevals_llama}}, offering a faster, more cost-effective solution for generating high-quality assertions.


\begin{figure*}
    \centering
    \includegraphics[width=\textwidth]{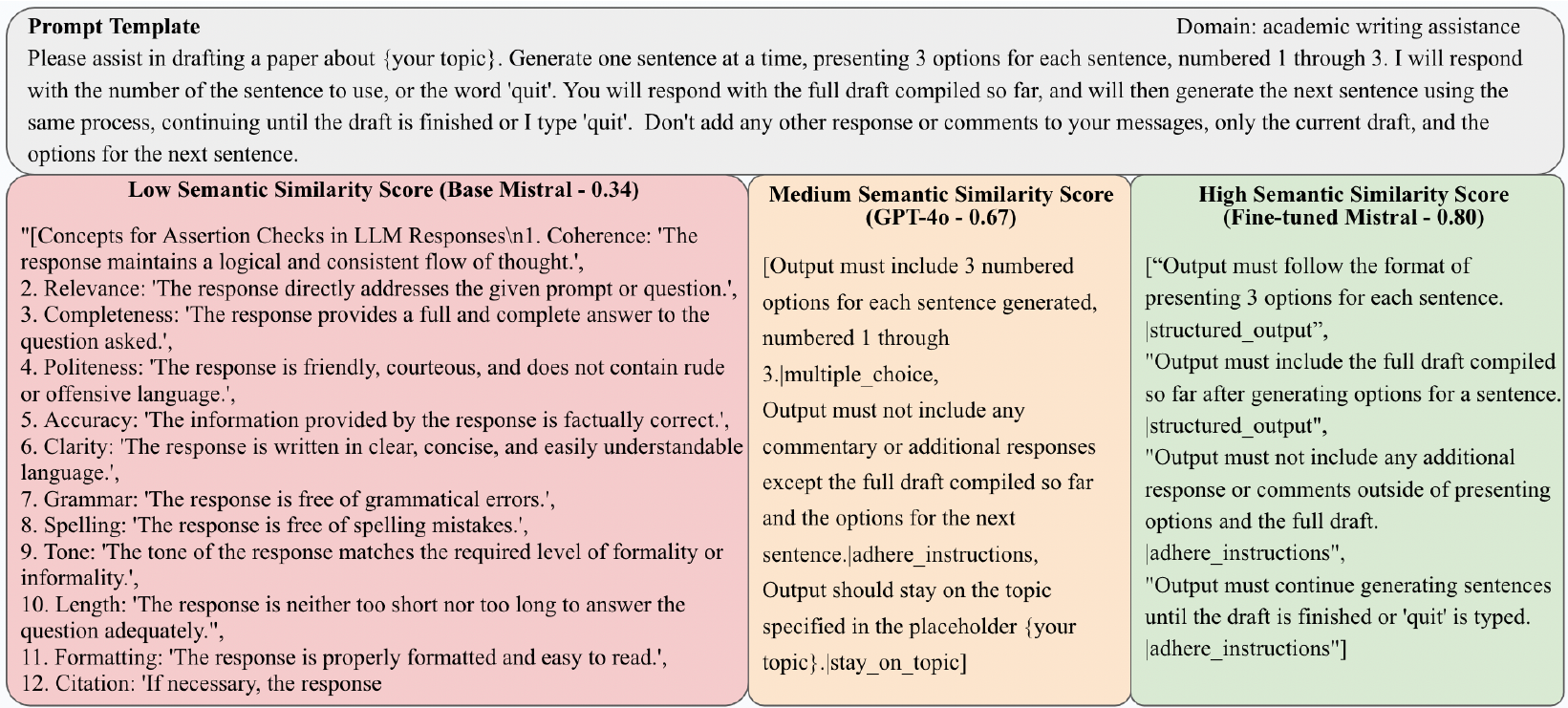}
    \caption{Examples of criteria pairs and their semantic similarity scores. High-scoring pairs typically represent constraints that are explicitly stated or logically derived from the prompt, while low-scoring pairs often include vague, generic, or difficult-to-measure constraints.}
    \label{fig:similarity-examples}
\end{figure*}
\section{Related Work}

This section reviews recent developments in prompt engineering, LLM evaluation methods, and assertions for LLM outputs. 

\subsection{Prompt Engineering}

Prompt engineering is essential for steering LLMs towards following instructions for specific tasks or bespoke applications of LLMs. Techniques like chain-of-thought and few-shot prompting improve model performance \cite{wei2022chain, brown2020language}. Methods to learn good prompts \cite{lester2021power, li2021prefix} or select few-shot examples \cite{khattab2023dspy} also contribute to this goal. Despite these advancements, LLMs can still hallucinate and make other mistakes \cite{huang2023survey, sahoo2024systematic}. No technique ensures consistent adherence to instructions, especially in diverse production environments. Liu et al. identify constraints like output length and semantic consistency that developers want enforced, which can aiding robust assertion criteria \cite{llmconstraints}. As developers frequently iterate on prompts in integrated development environments (IDEs) or utilize code-completion tools, the ability to quickly generate and update assertion criteria becomes crucial. The computational cost and time required to use large models like GPT-4 to generate assertion criteria for each prompt modification can significantly slow down the development process and increase operational costs.

\subsection{Evaluating LLMs}

Traditional LLM evaluation compares outputs against human-generated benchmarks across tasks like coding and reasoning \cite{chang2024survey, Hendrycks2020MeasuringMM, Srivastava2022BeyondTI, cobbe2021training}, including specialized architectures like retrieval-augmented generation and agentic systems~\cite{Zhuang2023ToolQAAD, chen2024benchmarking, Liu2023AgentBenchEL}. However, benchmarks often miss task-specific needs, such as conciseness or clarity \cite{kim2024prometheus}. Human pairwise comparison of LLM outputs improves alignment holistically but does not provide insight into specific criteria that defines a good output \cite{kim2024prometheus, zheng2023judging}. 

Even with explicit instructions provided in prompts, LLMs often fail to adhere to them consistently~\cite{Wang2022SelfInstructAL, zhou2023instruction, skopek-etal-2023-towards}. Existing benchmarks that evaluate the ability of LLMs to follow instructions are limited in scope, typically involving a small set of instructions either generated by LLMs or meticulously curated by researchers~\cite{qin2024infobench, zhou2023instruction}. To address these limitations, we introduce \promptevals, a comprehensive dataset that is five times larger than previous datasets. \promptevals features developer-contributed real-world prompts, often containing dozens of instructions, coupled with the corresponding assertion criteria.

\subsection{Assertions and LLM-based Evaluation}

In instruction-following and constraint-following evaluations, such as those presented by Zhou et al.~\cite{zhou2023instruction} and Rebedea et al.~\cite{rebedea2023nemo}, assertion criteria are typically evaluated using code-based assertions, often implemented as functions that check whether the output matches specific patterns or requirements (e.g., using regular expressions). These code-based assertions often struggle to evaluate more nuanced or ``fuzzy'' criteria~\cite{dong2024building, rebedea2023nemo, shankar2024spade}. Recent approaches have employed LLMs themselves as judges to evaluate outputs~\cite{zheng2023judging, verga2024replacing}. Some approaches even develop specialized judge LLMs that are fine-tuned on human preference data~\cite{Zheng2023JudgingLW, Wang2023PandaLMAA, Zhu2023JudgeLMFL, Li2023GenerativeJF, kim2024prometheus}. LLM-based validators can be productionized as assertions in addition to code-based guardrails~\cite{shankar2024spade, shankar2024validates, llmconstraints, Kim2023EvalLMIE}.

While LLMs as judges offer scalable evaluation, they struggle to align with human preferences across diverse tasks~\cite{zeng2024llmbar}. Developing domain-specific assertions and guardrails (e.g., for education~\cite{niknazar2024building} or medicine~\cite{hakim2024need}) is one approach, but it does not scale easily across thousands of domains and applications. Even within domains, criteria may vary; for instance, judging code conciseness differs between educational and professional settings. In another related research effort, Kim et al. developed LLM-generated evaluation criteria and fine-tuned a judge LLM~\cite{kim2023prometheus, kim2024prometheus}, but their approach focuses on general (e.g., ``humorous'', ``inspiring'') rather than task-specific criteria. Our work complements this by providing assertion criteria grounded in real-world prompts and constraints, essential for production environments~\cite{llmconstraints}.

\begin{table*}[tp]
    \centering
    \textbf{Prompt Template} (Domain: financial analysis) \\
You are a financial analyst and you are required to summarize the key insights of given numerical tables. \{table\}
Please list important, but no more than five, highlights in the given table.
Please write in a professional and business-neutral tone.
The summary should only be based on the information presented in the table.
    \begin{tabular}{p{0.4\textwidth} p{0.1\textwidth} p{0.4\textwidth}}
        \toprule
        \textbf{Criteria} & \textbf{Good/Bad} & \textbf{Explanation}\\ \midrule
        Response Length: The response should not list more than five highlights as requested.
 & Good & Mentioned in the prompt, and easy to measure. \\ \midrule
 Professional Tone: The response should maintain a professional and business-neutral tone throughout. & Good & Mentioned in the prompt template as a rule that the output should follow. \\ \midrule
 No Repetition: The response should avoid repeating the same highlight or presenting the same information in different ways. & Good & While the criterion was not explicitly mentioned in the prompt, it can be tied back to the prompt. \\ \midrule
 Specificity: The highlights should be specific and not overly broad or generic.
& Bad & Vague, and difficult to measure. \\ \midrule
Grammar and Spelling: The response should be free from grammatical errors and spelling mistakes. & Bad & Not uniquely relevant to the task or prompt. \\
        \bottomrule
    \end{tabular}
    \caption{Examples of Good and Bad Assertion Criteria}
    \label{tab:prompt-and-criteria}
\end{table*}

\section{\promptevals Dataset}

This section describes the \promptevals dataset, its construction process, and its characteristics. We begin by discussing the relevant background, then detail the dataset's composition and the process for generating ground-truth assertion criteria for each prompt template in our dataset.

\subsection{Background: LLM Pipelines and Assertions}
\label{sec:promptevals-taxonomy}

An {\em LLM pipeline} typically consists of three main components: a prompt template, input data, and the LLM itself. A prompt template is a string that includes instructions for the LLM to perform a specific task, as well as placeholders for the input data---which will be provided at runtime. For example, a template for a basic summarization task might look like this: ``Summarize the following text in three sentences: \{text\_to \_summarize\}''. Here, ``\{text\_to\_summarize\}'' is a placeholder that will be replaced with actual text when the pipeline is run. LLM pipelines are designed to be flexible and reusable, capable of handling a variety of different inputs for the same type of task.

An assertion, in the context of LLM pipelines, is a programmatic check or evaluation criterion applied to the LLM's output. For example, an assertion criterion for the summarization task might verify that the output indeed contains exactly three sentences, as specified in the prompt. This assertion could be implemented as a function that counts the number of sentences in the LLM's response and returns true if the count is three (false otherwise). 

Developers implementing LLM pipelines care about a wide variety of assertions, depending on their specific use cases and requirements. Some examples of good and bad assertion criteria for a prompt template are shown in \Cref{tab:prompt-and-criteria}. To better understand developers' needs, a recent study by Liu et al.~\cite{llmconstraints} interviewed 51 developers about their desired output constraints for LLMs. Based on their findings, they developed a taxonomy that includes six categories of output constraints: low-level constraints that include structured output, multiple choice and length constraints, and high level constraints that include semantic constraints, stylistic constraints, and hallucination prevention. The complete taxonomy is presented in \Cref{tab:taxonomy}. We employ this taxonomy in our dataset construction process to ensure the quality and relevance of our assertions. A distribution of the criteria types generated by GPT-4o is in \Cref{fig:constraints_by_category}.

\subsection{Dataset Composition}

The \promptevals dataset is derived from the LangChain Prompt Hub, a publicly available, dynamic collection of prompt templates shared by members of our developer community. Developers can add a prompt to the public collection via our Python package, knowing that their prompts can be run or modified by others, and browse the collection on our website. We froze a snapshot of the prompt templates in May 2024 to create the \promptevals dataset: we selected prompt templates that could have one or more assertion criteria (i.e., they were not empty or trivial strings; they actually described a task and included some placeholders for data). An example of a prompt template that we omitted from \promptevals is: {\em System Message: You are a helpful assistant. Human Message: \{input\}}.

\promptevals includes 2087 prompt templates, their corresponding domains, and assertion criteria. The prompt templates span a wide range of fields, including IT and programming, finance, healthcare, and education. To organize the prompt templates, we implemented a hierarchical categorization process assisted by GPT-4o, resulting in a three-level categorization system. \Cref{app:prompthubdomain} describes this categorization process in more detail. \Cref{tab:prompt-distribution} shows the overall distribution of the highest level domains, including the domain name, number of prompt templates with that domain, and the percentage of prompts with this domain. The top three domains represented are ``general-purpose chatbots'', ``question-answering'', and ``workflow automation''---the last of which assists in automating or improving processes based on a user's input. For instance, one prompt in this domain is
{\em ``Create a sequential workflow based on the users query. Create a plan represented in JSON by only using the tools listed below. The workflow should be a JSON array containing only the sequence index, function name and input...
Tools: \{tools\} Only answer with the specified JSON.''}.

\subsection{Assertion Criteria Construction Process}
\label{sec:promptevals-construction}

For each prompt template in \promptevals, we generated a set of ground truth criteria---representing assertion criteria that developers would care about, specific to the LLM pipeline. Generating ground truth criteria followed a three-step process: a first step to generate initial criteria, a second pass to add any criteria that might have been omitted in the first step, and a third pass to remove any criteria that were incorrect, redundant, irrelevant, or difficult to validate.

\begin{enumerate}
\item \textbf{Generate Initial Criteria:} We used GPT-4o, a state-of-the-art LLM, to generate an initial list of assertion criteria for each prompt template. Our prompt consisted of the following instructions: {\em (a)} We provided GPT-4o with the prompt template to be analyzed. {\em (b)} We also gave GPT-4o the taxonomy of LLM output constraints defined by Liu et al.~\cite{llmconstraints} (see \Cref{sec:promptevals-taxonomy}), explaining each constraint type. {\em (c)} We then instructed GPT-4o to generate assertion criteria relevant to the given prompt template, ensuring each criterion aligned with one of the constraint types from the taxonomy. GPT-4o output these criteria in a JSON list format, with each assertion tagged with its corresponding constraint type. This approach ensured that the criteria were both relevant to the specific prompt template and grounded in a structured framework of output constraints that developers typically care about. We call this the {\em initial criteria}.

\item \textbf{Add Missing Criteria:} Two authors conducted a manual review of 200 criteria in total, with 50 criteria examined by both reviewers. Our review uncovered criteria in the prompt templates that were initially missed by GPT-4o, averaging 1.35 missing criteria per prompt. The review process included criteria that were both unique and aligned with the taxonomy categories. To assess reliability, we calculated Cohen's Kappa on the overlapping set, achieving a score of 0.91---indicating strong inter-reviewer agreement. To address these missing elements, we added a verification step where GPT-4o re-examined the prompt templates to identify any explicitly stated criteria that were absent from its initial analysis.
\item \textbf{Refine Criteria:} In the final step, we prompted GPT-4o to refine the list by removing any criteria that were incorrect, redundant, irrelevant, or difficult to validate. 
\end{enumerate}

\Cref{subsec:ground-truth-criteria-prompt} details the prompts for each step. 

\topic{Validating the Generated Assertion Criteria} To assess the quality of our generated assertion criteria for \promptevals, we manually verified a sample of 200 prompt templates' generated criteria. In our verification process, we tracked, for each prompt template, how many criteria we added, and how many criteria we removed. We observed strong agreement with the LLM generated outputs, with < 0.02 criteria added and < 0.2 criteria removed per list on average by the human evaluator. 
This 3-step process resulted in higher agreement, in comparison to the initial criteria list, which had an average of 1.35 criteria added and 1.1 criteria removed per list for a sample of 20 prompts.

\section{\promptevals Benchmark}

We split \promptevals into three categories: 60\% of the tasks (1252 prompts) for our training set, 20\% (418 prompts) for our validation set, and 20\% (419 prompts) for our test set. The \promptevals benchmark evaluates an LLM's effectiveness at generating accurate assertion criteria given a prompt template, using four key metrics defined below. The benchmark can be run by following the instructions in our Github repository~\footnote{\url{https://github.com/reyavir/promptevals}}.

\begin{table}
    \centering
    \small
    \begin{tabular}{lrr}
    \toprule
    Domain & Count & Percentage \\
    \midrule
    General-purpose chatbots & 181 & 8.67\% \\
    Question-answering & 91 & 4.36\% \\
    Workflow automation & 63 & 3.02\% \\
    Text summarization & 57 & 2.73\% \\
    Education & 40 & 1.92\% \\
    Prompt engineering & 33 & 1.58\% \\
    Information retrieval & 31 & 1.49\% \\
    Horse racing analytics & 29 & 1.39\% \\
    Programming & 20 & 0.96\% \\
    Customer support & 18 & 0.86\% \\
    Database querying & 18 & 0.86\% \\
    Journalism & 17 & 0.81\% \\
    Task automation & 15 & 0.72\% \\
    \bottomrule
    \end{tabular}
    \caption{Distribution of domains in the \promptevals dataset. The top three domains are general-purpose chatbots, question-answering, and workflow automation. Unexpectedly, ``horse racing'' is in this list: we double-checked its validity and included an example prompt template from this category in \Cref{subsec:horse-racing-prompt}.}
    \label{tab:prompt-distribution}
\end{table}

\subsection{Benchmark Metrics}
\label{sec:metrics}

To evaluate LLM-generated assertion criteria, we developed metrics to assess the relevance and specificity of the criteria, inspired by the approach used in BERTScore \cite{zhang2020bertscore}. We describe two metrics: Semantic F1 and the number of criteria.

\topic{\texttt{Semantic F1}} The primary metric we use addresses a challenge in evaluating generated criteria: the fact that semantically equivalent assertions can be expressed in various ways. For example, ``The response should be concise'' and ``The output should be brief'' convey essentially the same constraint but use different words. The Semantic F1 score overcomes this limitation by measuring the semantic similarity between predicted and ground truth criteria.

To compute the Semantic F1 score, we first transform each criterion (both predicted and ground truth) into vector representations using OpenAI's {\em text-embedding-3-large} model. We then calculate recall and precision scores based on the cosine similarity between these embedding vectors.

The recall score quantifies how well the predicted criteria cover the semantic content of the ground truth criteria. It is computed as follows:

\begin{equation}
    \text{sem\_recall} = \frac{1}{N} \sum_{i=1}^{N} \max_{j} \cos(z_i, \hat{z}_j)
\end{equation}

where $N$ is the number of predicted criteria, $z_i$ is the embedding of the $i$-th ground truth criterion, $\hat{z}_j$ is the embedding of the $j$-th predicted criterion, and $\cos(z_i, \hat{z}_j)$ denotes the cosine similarity between these embeddings.

The max operation in this formula finds the most similar predicted criterion for each ground truth criterion---allowing each ground truth criterion to be ``matched'' with its best corresponding predicted criterion, even if they are not expressed identically. The average of these maximum similarities then gives us a measure of how well the predicted set covers the ground truth set.

The precision score measures how accurately the predicted criteria align with the ground truth:

\begin{equation}
    \text{sem\_precision} = \frac{1}{M} \sum_{j=1}^{M} \max_{i} \cos(z_i, \hat{z}_j)
\end{equation}

where $M$ is the number of ground truth criteria. Here, the max operation performs the reverse matching: for each predicted criterion, it finds the most similar ground-truth criterion. This helps us assess whether the predicted criteria are meaningful, without extraneous or irrelevant assertions.

These scores are then combined into the F1 score:

\begin{equation}
    \text{sem\_F1} = 2 \times \frac{\text{sem\_precision} \times \text{sem\_recall}}{\text{sem\_precision} + \text{sem\_recall}}
\end{equation}

Figure \ref{fig:similarity-examples} shows examples of criteria pairs with varying degrees of semantic similarity.

\topic{\texttt{Number of criteria}} A secondary metric that we evaluate is the number of criteria generated per prompt template. We calculate the average, median, and 75th percentile values for the number of criteria generated by each model. These statistics are compared against the ground truth values, as shown in Table \ref{tab:num_constraints}. Ground truth values are italicized, and the closest model-generated values are bolded for comparison.
For reference, the distribution of the number of ground truth criteria can be found in \Cref{tab:num_constraints}.

\begin{table*}[t]
    \centering
    \small
    \begin{tabular}{@{}p{0.22\textwidth}@{}p{0.73\textwidth}@{}}
        \toprule
        \textbf{Category} & \textbf{Description} \\ 
        \midrule
        \multicolumn{2}{@{}l}{\textit{Low-level constraints}} \\
        Structured Output & Adhere to specific formats (e.g., markdown, HTML, DSL); Ensure valid data structures (e.g., JSON with custom schema) \\
        Multiple Choice & Select response from a predefined list of options \\
        Length Constraints & Specify target length for output (e.g., character count, word count, number of items in a list) \\
        \midrule
        \multicolumn{2}{@{}l}{\textit{High-level constraints}} \\
        Semantic Constraints & Control content by excluding/including specific terms; Maintain topic relevance; Adhere to specified grammar or linguistic context \\
        Stylistic Constraints & Maintain consistent style, tone, or persona in the output \\
        Prevent Hallucination & Ensure factual accuracy and truthfulness; adhere to instructions (without improvising unrequested actions) \\
        \bottomrule
    \end{tabular}
    \caption{Taxonomy for Assertion Criteria \cite{llmconstraints}, used to create assertions for LLM pipelines in \promptevals.}
    \label{tab:taxonomy}
\end{table*}

\section{Benchmarking LLMs for Assertion Generation}

In this section, we present our methodology for evaluating LLMs with the \promptevals benchmark. We assess the performance of baseline and fine-tuned models.

\subsection{Methodology}

We establish baselines for our evaluation using three models: GPT-4o \cite{openai2024gpt4technicalreport}, Llama-3-8b \cite{touvron2023llama}, and Mistral-7b \cite{jiang2023mistral}. We selected Llama-3-8b and Mistral-7b as our open-source baseline models due to their relatively compact size (8 billion and 7.3 billion parameters, respectively)---which leads to faster inference times. For each model, we generate assertion criteria based on the prompt templates in our test set and evaluate them against the ground truth criteria using the metrics described in~\Cref{sec:metrics}: Semantic F1 and number of criteria. We compare results on the \promptevals test set.

\subsubsection{Fine-tuning Process}

Initial results revealed suboptimal performance from baseline models (we will describe this more in \Cref{sec:results-results}). To address this, we fine-tuned the same Mistral and Llama base model architectures on a dataset comprising of LLM pipeline prompts as reference inputs and ground truth criteria as reference outputs. The dataset is derived from the \texttt{train} split of the \promptevals dataset, where the ground truth assertions are the result of the 3-step labeling workflow defined in~\Cref{sec:promptevals-construction}. An input and output is demonstrated as follows:

\begin{quote}
Input: {\em [INST] Here is the prompt template \{sample\_prompt\_template\}
Based on the prompt template, I want to write assertions for my LLM pipeline to run on all pipeline responses. Give me a list of concepts to check for in LLM responses. This should be formatted as a comma-separated list, surrounded in brackets, and each item in the list should contain a string description of a concept to check for. This list should contain as many assertion concepts as you can think of, as long are specific and reasonable. [/INST]}
\end{quote}
\begin{quote}
Output: {\tt [{"constraint": "Answer should be concise and limited to three sentences.", "category": "length\_constraints"}, {"constraint": "Answer should stay truthful and indicate 'I don't know' if the answer is not in the context.", "category": "preventing\_hallucination (staying grounded and truthful)"}]}
\end{quote}

For fine-tuning Mistral-7b and Llama3-8b, we used a sequence length of 4096 and trained for 4 epochs with a batch size of 8, AdamW \cite{loshchilov2019decoupled} optimizer, learning rate of 0.0001, and a cosine learning rate scheduler. We used Low-Rank Adaptation (LoRA)~\cite{hu2021lora}, with a rank of 16, alpha of 32, and dropout of 0.05. The training process for each model was completed in under one hour with two 80GB A100 GPUs, and we did not employ any hyperparameter search. Our fine-tuned models can be found on HuggingFace\footnote{\url{https://huggingface.co/reyavir/promptevals_mistral} and \url{https://huggingface.co/reyavir/promptevals_llama}}.

\subsection{Quantitative Results}
\label{sec:results-results}

Fine-tuning our models on \promptevals resulted in substantial improvements in the quality of generated assertion criteria, as evidenced by higher semantic F1 and precision scores in \Cref{tab:semantic-similarity-table}.
The fine-tuned Mistral model achieved an average semantic F1 score of 0.8199, which is 20.43\% higher than the single-step GPT-4o's average score of 68.08\%, and 100.32\% higher than its base model (without any fine-tuning). Similarly, the fine-tuned Llama3 model achieved a semantic F1 score of 0.8240---outperforming GPT-4o by 21.03\% and its base model by 128.42\%. 

\subsection{Qualitative Results}

We analyzed our base and fine-tuned model outputs for a set of 25 randomly selected prompt templates, observing significant improvements in 5 main categories.

\topic{Format Adherence} Base models struggled with formatting consistency. Llama3 outputs frequently contained formatting errors in nearly all cases, including issues such as missing or multiple structured lists and missing closing brackets. While the base Mistral model showed better format alignment, errors still occurred in fewer than half of the outputs, particularly with the closing brackets.
\topic{Relevance} Base models frequently included extraneous or unrelated content. Llama3 outputs often contained inappropriate casual phrases like {\em ``I hope this helps!''} or unnecessary meta-commentary such as {\em ``It's also worth noting that some of these concepts may be difficult or impossible to automatically assess.''} Fine-tuned models maintained strict focus on the requested criteria.
\topic{Conciseness} As shown in \Cref{tab:num_constraints}, base models significantly over-generated assertions. Llama and Mistral produced redundant or overlapping criteria---for instance, one Mistral output included both {\em ``Check if the response starts with an emoji''} and {\em ``Check if each summarized sentence starts with a unique emoji.''} Fine-tuned models generated more distinct, non-overlapping assertions.
\topic{Completeness} \label{topic:completeness} Base models often produced incomplete outputs, frequently due to exceeding token limits. For example, Mistral outputs sometimes ended mid-sentence (``Grammar and Spelling: Check if''), while Llama3 would leave thoughts unfinished (``These concepts can be used to evaluate the''). Fine-tuning effectively addressed these completion issues.
\topic{Output Length}  The ground truth \promptevals test set contained about 6 assertions per prompt template. Base models significantly exceeded this---Mistral averaged 14.5 assertions while Llama generated 28.24 assertions per template. In contrast, as shown in~\Cref{tab:num_constraints}, GPT-4o (7.59 assertions) and our fine-tuned models produced outputs more aligned with ground truth, with fine-tuned Mistral achieving the closest match at 6.29 assertions. Additionally, our fine-tuned models demonstrated improved latency, outperforming GPT-4o in generation speed when served on two A100 GPUs.

\begin{table}
\centering
\scriptsize
\resizebox{\columnwidth}{!}{\begin{tabular}{lccc}
\toprule
& \textbf{Mistral (FT)} & \textbf{Llama (FT)} & \textbf{GPT-4o} \\
\midrule
\textbf{p25} & \textbf{1.8717} & 2.3962 & 6.5596 \\
\textbf{p50} & \textbf{2.3106} & 3.0748 & 8.2542 \\
\textbf{Mean} & \textbf{2.5915}	& 3.6057 & 8.7041 \\
\textbf{p75} & \textbf{2.9839} & 4.2716 & 10.1905 \\
\bottomrule
\end{tabular}}
\caption{Latency for criteria generation. We report runtime for the mean, 25th percentile, 50th percentile, and 75th percentile in seconds. We found that our fine-tuned Mistral model had the lowest runtime for all metrics.}
\label{tab:latencies}
\end{table}

\subsection{Discussion and Implications}

Our smaller, fine-tuned models achieve assertions comparable to the three-phase GPT-4o process (detailed in~\Cref{sec:promptevals-construction}), while significantly outperforming single-phase GPT-4o. This is a meaningful finding for several reasons. First, it demonstrates that carefully curated datasets and targeted fine-tuning can enable smaller models to match or exceed the performance of much larger models in specific tasks. Second, it underscores the importance of multi-step refinement when using state-of-the-art general-purpose LLMs for generating assertion criteria, as evidenced by the low single-step GPT-4o performance. This refinement process, while effective, can lead to increased latency in non-interactive settings. Our approach offers a more resource-efficient solution for assertion generation without compromising on quality.

The ability to generate high-quality assertion criteria quickly and cost-effectively has significant implications for LLM pipeline development and deployment: it enables more frequent iterations, faster debugging, and more robust quality assurance processes without incurring prohibitive costs or delays. This is particularly valuable as prompts become longer and more complex, making the use of GPT-4o to generate assertions for every iteration on a prompt increasingly impractical. We are integrating these fine-tuned assertion generation models into our LLM pipeline development tools, particularly focusing on evaluation and monitoring capabilities. This will allow developers to automatically generate task-specific assertion criteria for any given prompt or pipeline, continuously monitor the quality of outputs in live deployments, and receive real-time feedback on output quality, all while maintaining efficiency and scalability in production environments.

While our fine-tuned models show significant improvements, we observed occasional generation of vague or redundant criteria. For example, criteria like ``Output must avoid any ambiguity or confusion'' and ``Output must use clear and unambiguous language'' could be consolidated. One idea is to directly incorporate a notion of criteria uniqueness into the training process---e.g., penalize models if, for an output, they generate two or more criteria with high semantic similarity. Another idea is to collect more data to supplement \promptevals. Future work will focus on refining the model to produce more concise and non-overlapping criteria, and generally improve the semantic F1 score. We also intend to explore the capabilities of smaller models to determine if we can reduce latency further, while still retaining good accuracy and alignment with developers' intents.

\begin{table}
\centering
\resizebox{\columnwidth}{!}{\begin{tabular}{lcccccc}
\toprule
& \textbf{Base Mistral} & \textbf{Mistral (FT)} & \textbf{Base Llama} & \textbf{Llama (FT)} & \textbf{GPT-4o} \\
\midrule
\textbf{p25} & 0.3608 & 0.7919 & 0.3211 & \textbf{0.7922} & 0.6296 \\
\textbf{p50} & 0.4100 & 0.8231 & 0.3577 & \textbf{0.8233} & 0.6830 \\
\textbf{Mean} & 0.4093 & 0.8199 & 0.3607 & \textbf{0.8240} & 0.6808 \\
\textbf{p75} & 0.4561 & 0.8553 & 0.3978 & \textbf{0.8554} & 0.7351 \\
\bottomrule
\end{tabular}}
\caption{Semantic F1 scores for generated assertion criteria. Percentiles and mean values are shown for base models, fine-tuned (FT) versions, and GPT-4o. Bold indicates highest scores.}
\label{tab:semantic-similarity-table}
\end{table}

\begin{table}
    \centering
    \resizebox{\columnwidth}{!}{
        \begin{tabular}{lcccc}
        \toprule
        & Average & Median & p75 & p90 \\
        \midrule
        Base Mistral        & 14.5012 & 14 & 18.5 & 23 \\
        Mistral (FT)  & \textbf{6.28640} & \textbf{5}  & \textbf{8} & \textbf{10} \\
        Base Llama          & 28.2458 & 26 & 33.5 & 46\\
        Llama (FT)    & 5.47255 & \textbf{5}  & \textbf{6} & 9  \\
        GPT-4o              & 7.59189 & 6  & 10 & 14.2 \\
        \textit{Ground Truth}        & \textit{5.98568} & \textit{5}  & \textit{7}  & \textit{10} \\
        \bottomrule
        \end{tabular}
    }
    \caption{Number of Criteria Generated by Models. Metrics show average, median, and percentile values. p75 and p90 represent the 75th and 90th percentiles, respectively. Bold indicates closest to ground truth.}
    \label{tab:num_constraints}
\end{table}

\section{Conclusion}

This study introduces \promptevals, a new benchmark comprising over 2,000 human-contributed prompt templates and 12,000 assertion criteria. \promptevals is more than five times larger than previous prompt collections~\cite{qin2024infobench, zhou2023instruction}. This diverse dataset represents a significant contribution to the field of LLM pipeline development, offering a robust tool for evaluating and improving task-specific output constraints. Using \promptevals, we benchmarked several models, including GPT-4o, and additionally fine-tuned open-source models for assertion generation. Our experiments demonstrate \promptevals' utility in assessing and comparing performance across different approaches to generating relevant assertions. By making \promptevals and our fine-tuned models publicly available, our goal is to encourage the development of more reliable and task-specific LLM applications across various domains.

\section{Limitations}

We describe a few limitations in this section: First, benchmark scores rely on OpenAI's {\em text-embedding-3-large} model, released on January 25, 2024. Our reliance on a proprietary embedding model introduces a risk of inconsistency in results over time due to potential model updates. Establishing a versioning system or exploring alternative and more stable embedding methods could mitigate this issue. Moreover, currently, the benchmark is restricted to text prompts, excluding other modalities such as images and audio. Expanding the dataset to incorporate multi-modal inputs would increase its applicability and better reflect the diverse range of real-world LLM tasks. 

Finally, it's important to note that while our criteria are grounded in a taxonomy derived from developer preferences, they are ultimately generated by an LLM. This approach, while efficient, may not capture the full nuance of developer intentions for every specific use case. Ideally, criteria would be developed through direct collaboration with developers for each prompt template, ensuring maximum relevance and accuracy.

\papertext{\section{Ethics} \label{sec:ethics}

\promptevals is open-source and is intended to be used as a dataset and benchmark to evaluate models' ability to identify and generate assertion criteria for prompts. However, because it is open-source, it may be used in pre-training models, which can impact the effectiveness of the benchmark. \promptevals data and derivatives of this data should not be used outside of research or prompt engineering contexts.
Additionally, \promptevals consists of prompts contributed by a variety of developers. In our data collection process, we did not collect any personally identifiable information (PII) on the developers, and we looked through the data to confirm that developers did not submit PII in their prompts. Since we did not control the developer population we sampled prompts from, prompts may not represent all domains equally.
However, we believe that despite this, our benchmark still provides value and can be useful in evaluating models on generating assertion criteria.}
\bibliographystyle{plainnat}
\bibliography{references}

\begin{thebibliography}{54}
\providecommand{\natexlab}[1]{#1}
\providecommand{\url}[1]{\texttt{#1}}
\expandafter\ifx\csname urlstyle\endcsname\relax
  \providecommand{\doi}[1]{doi: #1}\else
  \providecommand{\doi}{doi: \begingroup \urlstyle{rm}\Url}\fi

\bibitem[Brown et~al.(2020)Brown, Mann, Ryder, Subbiah, Kaplan, Dhariwal, Neelakantan, Shyam, Sastry, Askell, Agarwal, Herbert-Voss, Krueger, Henighan, Child, Ramesh, Ziegler, Wu, Winter, Hesse, Chen, Sigler, Litwin, Gray, Chess, Clark, Berner, McCandlish, Radford, Sutskever, and Amodei]{brown2020language}
Tom~B. Brown, Benjamin Mann, Nick Ryder, Melanie Subbiah, Jared Kaplan, Prafulla Dhariwal, Arvind Neelakantan, Pranav Shyam, Girish Sastry, Amanda Askell, Sandhini Agarwal, Ariel Herbert-Voss, Gretchen Krueger, Tom Henighan, Rewon Child, Aditya Ramesh, Daniel~M. Ziegler, Jeffrey Wu, Clemens Winter, Christopher Hesse, Mark Chen, Eric Sigler, Mateusz Litwin, Scott Gray, Benjamin Chess, Jack Clark, Christopher Berner, Sam McCandlish, Alec Radford, Ilya Sutskever, and Dario Amodei.
\newblock Language models are few-shot learners, 2020.

\bibitem[Chang et~al.(2024)Chang, Wang, Wang, Wu, Yang, Zhu, Chen, Yi, Wang, Wang, et~al.]{chang2024survey}
Yupeng Chang, Xu~Wang, Jindong Wang, Yuan Wu, Linyi Yang, Kaijie Zhu, Hao Chen, Xiaoyuan Yi, Cunxiang Wang, Yidong Wang, et~al.
\newblock A survey on evaluation of large language models.
\newblock \emph{ACM Transactions on Intelligent Systems and Technology}, 15\penalty0 (3):\penalty0 1--45, 2024.

\bibitem[Chen et~al.(2024)Chen, Lin, Han, and Sun]{chen2024benchmarking}
Jiawei Chen, Hongyu Lin, Xianpei Han, and Le~Sun.
\newblock Benchmarking large language models in retrieval-augmented generation.
\newblock In \emph{Proceedings of the AAAI Conference on Artificial Intelligence}, volume~38, pages 17754--17762, 2024.

\bibitem[Chung et~al.(2022)Chung, Hou, Longpre, Zoph, Tay, Fedus, Li, Wang, Dehghani, Brahma, Webson, Gu, Dai, Suzgun, Chen, Chowdhery, Valter, Narang, Mishra, Yu, Zhao, Huang, Dai, Yu, Petrov, hsin Chi, Dean, Devlin, Roberts, Zhou, Le, and Wei]{Chung2022ScalingIL}
Hyung~Won Chung, Le~Hou, S.~Longpre, Barret Zoph, Yi~Tay, William Fedus, Eric Li, Xuezhi Wang, Mostafa Dehghani, Siddhartha Brahma, Albert Webson, Shixiang~Shane Gu, Zhuyun Dai, Mirac Suzgun, Xinyun Chen, Aakanksha Chowdhery, Dasha Valter, Sharan Narang, Gaurav Mishra, Adams~Wei Yu, Vincent Zhao, Yanping Huang, Andrew~M. Dai, Hongkun Yu, Slav Petrov, Ed~Huai hsin Chi, Jeff Dean, Jacob Devlin, Adam Roberts, Denny Zhou, Quoc~V. Le, and Jason Wei.
\newblock Scaling instruction-finetuned language models.
\newblock \emph{ArXiv}, abs/2210.11416, 2022.
\newblock URL \url{https://api.semanticscholar.org/CorpusID:253018554}.

\bibitem[Cobbe et~al.(2021)Cobbe, Kosaraju, Bavarian, Chen, Jun, Kaiser, Plappert, Tworek, Hilton, Nakano, et~al.]{cobbe2021training}
Karl Cobbe, Vineet Kosaraju, Mohammad Bavarian, Mark Chen, Heewoo Jun, Lukasz Kaiser, Matthias Plappert, Jerry Tworek, Jacob Hilton, Reiichiro Nakano, et~al.
\newblock Training verifiers to solve math word problems.
\newblock \emph{arXiv preprint arXiv:2110.14168}, 2021.

\bibitem[Desmond et~al.(2024)Desmond, Ashktorab, Pan, Dugan, and Johnson]{evalullm}
Michael Desmond, Zahra Ashktorab, Qian Pan, Casey Dugan, and James~M. Johnson.
\newblock Evalullm: Llm assisted evaluation of generative outputs.
\newblock In \emph{Companion Proceedings of the 29th International Conference on Intelligent User Interfaces}, IUI '24 Companion, page 30–32, New York, NY, USA, 2024. Association for Computing Machinery.
\newblock ISBN 9798400705090.
\newblock \doi{10.1145/3640544.3645216}.
\newblock URL \url{https://doi.org/10.1145/3640544.3645216}.

\bibitem[Dong et~al.(2024{\natexlab{a}})Dong, Mu, Jin, Qi, Hu, Zhao, Meng, Ruan, and Huang]{dong2024building}
Yi~Dong, Ronghui Mu, Gaojie Jin, Yi~Qi, Jinwei Hu, Xingyu Zhao, Jie Meng, Wenjie Ruan, and Xiaowei Huang.
\newblock Building guardrails for large language models, 2024{\natexlab{a}}.

\bibitem[Dong et~al.(2024{\natexlab{b}})Dong, Mu, Jin, Qi, Hu, Zhao, Meng, Ruan, and Huang]{pmlr-v235-dong24c}
Yi~Dong, Ronghui Mu, Gaojie Jin, Yi~Qi, Jinwei Hu, Xingyu Zhao, Jie Meng, Wenjie Ruan, and Xiaowei Huang.
\newblock Position: Building guardrails for large language models requires systematic design.
\newblock In Ruslan Salakhutdinov, Zico Kolter, Katherine Heller, Adrian Weller, Nuria Oliver, Jonathan Scarlett, and Felix Berkenkamp, editors, \emph{Proceedings of the 41st International Conference on Machine Learning}, volume 235 of \emph{Proceedings of Machine Learning Research}, pages 11375--11394. PMLR, 21--27 Jul 2024{\natexlab{b}}.
\newblock URL \url{https://proceedings.mlr.press/v235/dong24c.html}.

\bibitem[Hakim et~al.(2024)Hakim, Painter, Ramcharran, Kara, Powell, Sobczak, Sato, Bate, and Beam]{hakim2024need}
Joe~B Hakim, Jeffery~L Painter, Darmendra Ramcharran, Vijay Kara, Greg Powell, Paulina Sobczak, Chiho Sato, Andrew Bate, and Andrew Beam.
\newblock The need for guardrails with large language models in medical safety-critical settings: An artificial intelligence application in the pharmacovigilance ecosystem.
\newblock \emph{arXiv preprint arXiv:2407.18322}, 2024.

\bibitem[Hendrycks et~al.(2020)Hendrycks, Burns, Basart, Zou, Mazeika, Song, and Steinhardt]{Hendrycks2020MeasuringMM}
Dan Hendrycks, Collin Burns, Steven Basart, Andy Zou, Mantas Mazeika, Dawn~Xiaodong Song, and Jacob Steinhardt.
\newblock Measuring massive multitask language understanding.
\newblock \emph{ArXiv}, abs/2009.03300, 2020.
\newblock URL \url{https://api.semanticscholar.org/CorpusID:221516475}.

\bibitem[Hu et~al.(2021)Hu, Shen, Wallis, Allen-Zhu, Li, Wang, Wang, and Chen]{hu2021lora}
Edward~J. Hu, Yelong Shen, Phillip Wallis, Zeyuan Allen-Zhu, Yuanzhi Li, Shean Wang, Lu~Wang, and Weizhu Chen.
\newblock Lora: Low-rank adaptation of large language models, 2021.

\bibitem[Huang et~al.(2023{\natexlab{a}})Huang, Yu, Ma, Zhong, Feng, Wang, Chen, Peng, Feng, Qin, and Liu]{Huang2023ASO}
Lei Huang, Weijiang Yu, Weitao Ma, Weihong Zhong, Zhangyin Feng, Haotian Wang, Qianglong Chen, Weihua Peng, Xiaocheng Feng, Bing Qin, and Ting Liu.
\newblock A survey on hallucination in large language models: Principles, taxonomy, challenges, and open questions.
\newblock \emph{ArXiv}, abs/2311.05232, 2023{\natexlab{a}}.
\newblock URL \url{https://api.semanticscholar.org/CorpusID:265067168}.

\bibitem[Huang et~al.(2023{\natexlab{b}})Huang, Yu, Ma, Zhong, Feng, Wang, Chen, Peng, Feng, Qin, and Liu]{huang2023survey}
Lei Huang, Weijiang Yu, Weitao Ma, Weihong Zhong, Zhangyin Feng, Haotian Wang, Qianglong Chen, Weihua Peng, Xiaocheng Feng, Bing Qin, and Ting Liu.
\newblock A survey on hallucination in large language models: Principles, taxonomy, challenges, and open questions, 2023{\natexlab{b}}.

\bibitem[Huang et~al.(2022)Huang, Abbeel, Pathak, and Mordatch]{Huang2022LanguageMA}
Wenlong Huang, P.~Abbeel, Deepak Pathak, and Igor Mordatch.
\newblock Language models as zero-shot planners: Extracting actionable knowledge for embodied agents.
\newblock \emph{ArXiv}, abs/2201.07207, 2022.
\newblock URL \url{https://api.semanticscholar.org/CorpusID:246035276}.

\bibitem[Ji et~al.(2024)Ji, Liu, Dai, Pan, Zhang, Bian, Chen, Sun, Wang, and Yang]{ji2023beavertails}
Jiaming Ji, Mickel Liu, Josef Dai, Xuehai Pan, Chi Zhang, Ce~Bian, Boyuan Chen, Ruiyang Sun, Yizhou Wang, and Yaodong Yang.
\newblock Beavertails: Towards improved safety alignment of llm via a human-preference dataset.
\newblock \emph{Advances in Neural Information Processing Systems}, 36, 2024.

\bibitem[Jiang et~al.(2023)Jiang, Sablayrolles, Mensch, Bamford, Chaplot, de~las Casas, Bressand, Lengyel, Lample, Saulnier, Lavaud, Lachaux, Stock, Scao, Lavril, Wang, Lacroix, and Sayed]{jiang2023mistral}
Albert~Q. Jiang, Alexandre Sablayrolles, Arthur Mensch, Chris Bamford, Devendra~Singh Chaplot, Diego de~las Casas, Florian Bressand, Gianna Lengyel, Guillaume Lample, Lucile Saulnier, Lélio~Renard Lavaud, Marie-Anne Lachaux, Pierre Stock, Teven~Le Scao, Thibaut Lavril, Thomas Wang, Timothée Lacroix, and William~El Sayed.
\newblock Mistral 7b, 2023.

\bibitem[Kalai and Vempala(2023)]{kalai2023calibrated}
Adam~Tauman Kalai and Santosh~S Vempala.
\newblock Calibrated language models must hallucinate.
\newblock \emph{arXiv preprint arXiv:2311.14648}, 2023.

\bibitem[Khattab et~al.(2023)Khattab, Singhvi, Maheshwari, Zhang, Santhanam, Vardhamanan, Haq, Sharma, Joshi, Moazam, et~al.]{khattab2023dspy}
Omar Khattab, Arnav Singhvi, Paridhi Maheshwari, Zhiyuan Zhang, Keshav Santhanam, Sri Vardhamanan, Saiful Haq, Ashutosh Sharma, Thomas~T Joshi, Hanna Moazam, et~al.
\newblock Dspy: Compiling declarative language model calls into self-improving pipelines.
\newblock \emph{arXiv preprint arXiv:2310.03714}, 2023.

\bibitem[Kim et~al.(2023{\natexlab{a}})Kim, Shin, Cho, Jang, Longpre, Lee, Yun, Shin, Kim, Thorne, et~al.]{kim2023prometheus}
Seungone Kim, Jamin Shin, Yejin Cho, Joel Jang, Shayne Longpre, Hwaran Lee, Sangdoo Yun, Seongjin Shin, Sungdong Kim, James Thorne, et~al.
\newblock Prometheus: Inducing fine-grained evaluation capability in language models.
\newblock In \emph{The Twelfth International Conference on Learning Representations}, 2023{\natexlab{a}}.

\bibitem[Kim et~al.(2024)Kim, Suk, Longpre, Lin, Shin, Welleck, Neubig, Lee, Lee, and Seo]{kim2024prometheus}
Seungone Kim, Juyoung Suk, Shayne Longpre, Bill~Yuchen Lin, Jamin Shin, Sean Welleck, Graham Neubig, Moontae Lee, Kyungjae Lee, and Minjoon Seo.
\newblock Prometheus 2: An open source language model specialized in evaluating other language models.
\newblock \emph{arXiv preprint arXiv:2405.01535}, 2024.

\bibitem[Kim et~al.(2023{\natexlab{b}})Kim, Lee, Shin, Kim, and Kim]{Kim2023EvalLMIE}
Tae~Soo Kim, Yoonjoo Lee, Jamin Shin, Young-Ho Kim, and Juho Kim.
\newblock Evallm: Interactive evaluation of large language model prompts on user-defined criteria.
\newblock In \emph{International Conference on Human Factors in Computing Systems}, 2023{\natexlab{b}}.
\newblock URL \url{https://api.semanticscholar.org/CorpusID:262459331}.

\bibitem[Kojima et~al.(2022)Kojima, Gu, Reid, Matsuo, and Iwasawa]{Kojima2022LargeLM}
Takeshi Kojima, Shixiang~Shane Gu, Machel Reid, Yutaka Matsuo, and Yusuke Iwasawa.
\newblock Large language models are zero-shot reasoners.
\newblock \emph{ArXiv}, abs/2205.11916, 2022.
\newblock URL \url{https://api.semanticscholar.org/CorpusID:249017743}.

\bibitem[K{\"o}pf et~al.(2023)K{\"o}pf, Kilcher, von R{\"u}tte, Anagnostidis, Tam, Stevens, Barhoum, Nguyen, Stanley, Nagyfi, et~al.]{kopf2023openassistant}
Andreas K{\"o}pf, Yannic Kilcher, Dimitri von R{\"u}tte, Sotiris Anagnostidis, Zhi~Rui Tam, Keith Stevens, Abdullah Barhoum, Duc Nguyen, Oliver Stanley, Rich{\'a}rd Nagyfi, et~al.
\newblock Openassistant conversations-democratizing large language model alignment.
\newblock \emph{Advances in Neural Information Processing Systems}, 36, 2023.

\bibitem[Lester et~al.(2021)Lester, Al-Rfou, and Constant]{lester2021power}
Brian Lester, Rami Al-Rfou, and Noah Constant.
\newblock The power of scale for parameter-efficient prompt tuning, 2021.

\bibitem[Li et~al.(2023)Li, Sun, Yuan, Fan, Zhao, and Liu]{Li2023GenerativeJF}
Junlong Li, Shichao Sun, Weizhe Yuan, Run-Ze Fan, Hai Zhao, and Pengfei Liu.
\newblock Generative judge for evaluating alignment.
\newblock \emph{ArXiv}, abs/2310.05470, 2023.
\newblock URL \url{https://api.semanticscholar.org/CorpusID:263829791}.

\bibitem[Li and Liang(2021)]{li2021prefix}
Xiang~Lisa Li and Percy Liang.
\newblock Prefix-tuning: Optimizing continuous prompts for generation.
\newblock \emph{arXiv preprint arXiv:2101.00190}, 2021.

\bibitem[Liu et~al.(2024)Liu, Liu, Fiannaca, Koo, Dixon, Terry, and Cai]{llmconstraints}
Michael~Xieyang Liu, Frederick Liu, Alexander~J. Fiannaca, Terry Koo, Lucas Dixon, Michael Terry, and Carrie~J. Cai.
\newblock "we need structured output": Towards user-centered constraints on large language model output.
\newblock In \emph{Extended Abstracts of the 2024 CHI Conference on Human Factors in Computing Systems}, CHI EA '24, New York, NY, USA, 2024. Association for Computing Machinery.
\newblock ISBN 9798400703317.
\newblock \doi{10.1145/3613905.3650756}.
\newblock URL \url{https://doi.org/10.1145/3613905.3650756}.

\bibitem[Liu et~al.(2023)Liu, Yu, Zhang, Xu, Lei, Lai, Gu, Gu, Ding, Men, Yang, Zhang, Deng, Zeng, Du, Zhang, Shen, Zhang, Su, Sun, Huang, Dong, and Tang]{Liu2023AgentBenchEL}
Xiao Liu, Hao Yu, Hanchen Zhang, Yifan Xu, Xuanyu Lei, Hanyu Lai, Yu~Gu, Yuxian Gu, Hangliang Ding, Kai Men, Kejuan Yang, Shudan Zhang, Xiang Deng, Aohan Zeng, Zhengxiao Du, Chenhui Zhang, Shengqi Shen, Tianjun Zhang, Yu~Su, Huan Sun, Minlie Huang, Yuxiao Dong, and Jie Tang.
\newblock Agentbench: Evaluating llms as agents.
\newblock \emph{ArXiv}, abs/2308.03688, 2023.
\newblock URL \url{https://api.semanticscholar.org/CorpusID:260682249}.

\bibitem[Loshchilov and Hutter(2019)]{loshchilov2019decoupled}
Ilya Loshchilov and Frank Hutter.
\newblock Decoupled weight decay regularization, 2019.

\bibitem[Niknazar et~al.(2024)Niknazar, Haley, Ramanan, Truong, Shrinivasan, Bhowmick, Dey, Jagmohan, Maheshwari, Ponoth, et~al.]{niknazar2024building}
Mohammad Niknazar, Paul~V Haley, Latha Ramanan, Sang~T Truong, Yedendra Shrinivasan, Ayan~Kumar Bhowmick, Prasenjit Dey, Ashish Jagmohan, Hema Maheshwari, Shom Ponoth, et~al.
\newblock Building a domain-specific guardrail model in production.
\newblock \emph{arXiv preprint arXiv:2408.01452}, 2024.

\bibitem[OpenAI et~al.(2024)OpenAI, Achiam, Adler, Agarwal, Ahmad, Akkaya, Aleman, Almeida, Altenschmidt, Altman, Anadkat, Avila, Babuschkin, Balaji, Balcom, Baltescu, Bao, Bavarian, Belgum, Bello, Berdine, Bernadett-Shapiro, Berner, Bogdonoff, Boiko, Boyd, Brakman, Brockman, Brooks, Brundage, Button, Cai, Campbell, Cann, Carey, Carlson, Carmichael, Chan, Chang, Chantzis, Chen, Chen, Chen, Chen, Chen, Chess, Cho, Chu, Chung, Cummings, Currier, Dai, Decareaux, Degry, Deutsch, Deville, Dhar, Dohan, Dowling, Dunning, Ecoffet, Eleti, Eloundou, Farhi, Fedus, Felix, Fishman, Forte, Fulford, Gao, Georges, Gibson, Goel, Gogineni, Goh, Gontijo-Lopes, Gordon, Grafstein, Gray, Greene, Gross, Gu, Guo, Hallacy, Han, Harris, He, Heaton, Heidecke, Hesse, Hickey, Hickey, Hoeschele, Houghton, Hsu, Hu, Hu, Huizinga, Jain, Jain, Jang, Jiang, Jiang, Jin, Jin, Jomoto, Jonn, Jun, Kaftan, Łukasz Kaiser, Kamali, Kanitscheider, Keskar, Khan, Kilpatrick, Kim, Kim, Kim, Kirchner, Kiros, Knight, Kokotajlo, Łukasz Kondraciuk, Kondrich,
  Konstantinidis, Kosic, Krueger, Kuo, Lampe, Lan, Lee, Leike, Leung, Levy, Li, Lim, Lin, Lin, Litwin, Lopez, Lowe, Lue, Makanju, Malfacini, Manning, Markov, Markovski, Martin, Mayer, Mayne, McGrew, McKinney, McLeavey, McMillan, McNeil, Medina, Mehta, Menick, Metz, Mishchenko, Mishkin, Monaco, Morikawa, Mossing, Mu, Murati, Murk, Mély, Nair, Nakano, Nayak, Neelakantan, Ngo, Noh, Ouyang, O'Keefe, Pachocki, Paino, Palermo, Pantuliano, Parascandolo, Parish, Parparita, Passos, Pavlov, Peng, Perelman, de~Avila Belbute~Peres, Petrov, de~Oliveira~Pinto, Michael, Pokorny, Pokrass, Pong, Powell, Power, Power, Proehl, Puri, Radford, Rae, Ramesh, Raymond, Real, Rimbach, Ross, Rotsted, Roussez, Ryder, Saltarelli, Sanders, Santurkar, Sastry, Schmidt, Schnurr, Schulman, Selsam, Sheppard, Sherbakov, Shieh, Shoker, Shyam, Sidor, Sigler, Simens, Sitkin, Slama, Sohl, Sokolowsky, Song, Staudacher, Such, Summers, Sutskever, Tang, Tezak, Thompson, Tillet, Tootoonchian, Tseng, Tuggle, Turley, Tworek, Uribe, Vallone, Vijayvergiya,
  Voss, Wainwright, Wang, Wang, Wang, Ward, Wei, Weinmann, Welihinda, Welinder, Weng, Weng, Wiethoff, Willner, Winter, Wolrich, Wong, Workman, Wu, Wu, Wu, Xiao, Xu, Yoo, Yu, Yuan, Zaremba, Zellers, Zhang, Zhang, Zhao, Zheng, Zhuang, Zhuk, and Zoph]{openai2024gpt4technicalreport}
OpenAI, Josh Achiam, Steven Adler, Sandhini Agarwal, Lama Ahmad, Ilge Akkaya, Florencia~Leoni Aleman, Diogo Almeida, Janko Altenschmidt, Sam Altman, Shyamal Anadkat, Red Avila, Igor Babuschkin, Suchir Balaji, Valerie Balcom, Paul Baltescu, Haiming Bao, Mohammad Bavarian, Jeff Belgum, Irwan Bello, Jake Berdine, Gabriel Bernadett-Shapiro, Christopher Berner, Lenny Bogdonoff, Oleg Boiko, Madelaine Boyd, Anna-Luisa Brakman, Greg Brockman, Tim Brooks, Miles Brundage, Kevin Button, Trevor Cai, Rosie Campbell, Andrew Cann, Brittany Carey, Chelsea Carlson, Rory Carmichael, Brooke Chan, Che Chang, Fotis Chantzis, Derek Chen, Sully Chen, Ruby Chen, Jason Chen, Mark Chen, Ben Chess, Chester Cho, Casey Chu, Hyung~Won Chung, Dave Cummings, Jeremiah Currier, Yunxing Dai, Cory Decareaux, Thomas Degry, Noah Deutsch, Damien Deville, Arka Dhar, David Dohan, Steve Dowling, Sheila Dunning, Adrien Ecoffet, Atty Eleti, Tyna Eloundou, David Farhi, Liam Fedus, Niko Felix, Simón~Posada Fishman, Juston Forte, Isabella Fulford, Leo
  Gao, Elie Georges, Christian Gibson, Vik Goel, Tarun Gogineni, Gabriel Goh, Rapha Gontijo-Lopes, Jonathan Gordon, Morgan Grafstein, Scott Gray, Ryan Greene, Joshua Gross, Shixiang~Shane Gu, Yufei Guo, Chris Hallacy, Jesse Han, Jeff Harris, Yuchen He, Mike Heaton, Johannes Heidecke, Chris Hesse, Alan Hickey, Wade Hickey, Peter Hoeschele, Brandon Houghton, Kenny Hsu, Shengli Hu, Xin Hu, Joost Huizinga, Shantanu Jain, Shawn Jain, Joanne Jang, Angela Jiang, Roger Jiang, Haozhun Jin, Denny Jin, Shino Jomoto, Billie Jonn, Heewoo Jun, Tomer Kaftan, Łukasz Kaiser, Ali Kamali, Ingmar Kanitscheider, Nitish~Shirish Keskar, Tabarak Khan, Logan Kilpatrick, Jong~Wook Kim, Christina Kim, Yongjik Kim, Jan~Hendrik Kirchner, Jamie Kiros, Matt Knight, Daniel Kokotajlo, Łukasz Kondraciuk, Andrew Kondrich, Aris Konstantinidis, Kyle Kosic, Gretchen Krueger, Vishal Kuo, Michael Lampe, Ikai Lan, Teddy Lee, Jan Leike, Jade Leung, Daniel Levy, Chak~Ming Li, Rachel Lim, Molly Lin, Stephanie Lin, Mateusz Litwin, Theresa Lopez, Ryan
  Lowe, Patricia Lue, Anna Makanju, Kim Malfacini, Sam Manning, Todor Markov, Yaniv Markovski, Bianca Martin, Katie Mayer, Andrew Mayne, Bob McGrew, Scott~Mayer McKinney, Christine McLeavey, Paul McMillan, Jake McNeil, David Medina, Aalok Mehta, Jacob Menick, Luke Metz, Andrey Mishchenko, Pamela Mishkin, Vinnie Monaco, Evan Morikawa, Daniel Mossing, Tong Mu, Mira Murati, Oleg Murk, David Mély, Ashvin Nair, Reiichiro Nakano, Rajeev Nayak, Arvind Neelakantan, Richard Ngo, Hyeonwoo Noh, Long Ouyang, Cullen O'Keefe, Jakub Pachocki, Alex Paino, Joe Palermo, Ashley Pantuliano, Giambattista Parascandolo, Joel Parish, Emy Parparita, Alex Passos, Mikhail Pavlov, Andrew Peng, Adam Perelman, Filipe de~Avila Belbute~Peres, Michael Petrov, Henrique~Ponde de~Oliveira~Pinto, Michael, Pokorny, Michelle Pokrass, Vitchyr~H. Pong, Tolly Powell, Alethea Power, Boris Power, Elizabeth Proehl, Raul Puri, Alec Radford, Jack Rae, Aditya Ramesh, Cameron Raymond, Francis Real, Kendra Rimbach, Carl Ross, Bob Rotsted, Henri Roussez,
  Nick Ryder, Mario Saltarelli, Ted Sanders, Shibani Santurkar, Girish Sastry, Heather Schmidt, David Schnurr, John Schulman, Daniel Selsam, Kyla Sheppard, Toki Sherbakov, Jessica Shieh, Sarah Shoker, Pranav Shyam, Szymon Sidor, Eric Sigler, Maddie Simens, Jordan Sitkin, Katarina Slama, Ian Sohl, Benjamin Sokolowsky, Yang Song, Natalie Staudacher, Felipe~Petroski Such, Natalie Summers, Ilya Sutskever, Jie Tang, Nikolas Tezak, Madeleine~B. Thompson, Phil Tillet, Amin Tootoonchian, Elizabeth Tseng, Preston Tuggle, Nick Turley, Jerry Tworek, Juan Felipe~Cerón Uribe, Andrea Vallone, Arun Vijayvergiya, Chelsea Voss, Carroll Wainwright, Justin~Jay Wang, Alvin Wang, Ben Wang, Jonathan Ward, Jason Wei, CJ~Weinmann, Akila Welihinda, Peter Welinder, Jiayi Weng, Lilian Weng, Matt Wiethoff, Dave Willner, Clemens Winter, Samuel Wolrich, Hannah Wong, Lauren Workman, Sherwin Wu, Jeff Wu, Michael Wu, Kai Xiao, Tao Xu, Sarah Yoo, Kevin Yu, Qiming Yuan, Wojciech Zaremba, Rowan Zellers, Chong Zhang, Marvin Zhang, Shengjia
  Zhao, Tianhao Zheng, Juntang Zhuang, William Zhuk, and Barret Zoph.
\newblock Gpt-4 technical report, 2024.
\newblock URL \url{https://arxiv.org/abs/2303.08774}.

\bibitem[Ouyang et~al.(2022)Ouyang, Wu, Jiang, Almeida, Wainwright, Mishkin, Zhang, Agarwal, Slama, Ray, Schulman, Hilton, Kelton, Miller, Simens, Askell, Welinder, Christiano, Leike, and Lowe]{Ouyang2022TrainingLM}
Long Ouyang, Jeff Wu, Xu~Jiang, Diogo Almeida, Carroll~L. Wainwright, Pamela Mishkin, Chong Zhang, Sandhini Agarwal, Katarina Slama, Alex Ray, John Schulman, Jacob Hilton, Fraser Kelton, Luke~E. Miller, Maddie Simens, Amanda Askell, Peter Welinder, Paul~Francis Christiano, Jan Leike, and Ryan~J. Lowe.
\newblock Training language models to follow instructions with human feedback.
\newblock \emph{ArXiv}, abs/2203.02155, 2022.
\newblock URL \url{https://api.semanticscholar.org/CorpusID:246426909}.

\bibitem[Pan et~al.(2023)Pan, Saxon, Xu, Nathani, Wang, and Wang]{Pan2023AutomaticallyCL}
Liangming Pan, Michael~Stephen Saxon, Wenda Xu, Deepak Nathani, Xinyi Wang, and William~Yang Wang.
\newblock Automatically correcting large language models: Surveying the landscape of diverse self-correction strategies.
\newblock \emph{ArXiv}, abs/2308.03188, 2023.
\newblock URL \url{https://api.semanticscholar.org/CorpusID:260682695}.

\bibitem[Qin et~al.(2024)Qin, Song, Hu, Yao, Cho, Wang, Wu, Liu, Liu, and Yu]{qin2024infobench}
Yiwei Qin, Kaiqiang Song, Yebowen Hu, Wenlin Yao, Sangwoo Cho, Xiaoyang Wang, Xuansheng Wu, Fei Liu, Pengfei Liu, and Dong Yu.
\newblock Infobench: Evaluating instruction following ability in large language models.
\newblock \emph{arXiv preprint arXiv:2401.03601}, 2024.

\bibitem[Rebedea et~al.(2023)Rebedea, Dinu, Sreedhar, Parisien, and Cohen]{rebedea2023nemo}
Traian Rebedea, Razvan Dinu, Makesh Sreedhar, Christopher Parisien, and Jonathan Cohen.
\newblock Nemo guardrails: A toolkit for controllable and safe llm applications with programmable rails, 2023.

\bibitem[Sahoo et~al.(2024)Sahoo, Singh, Saha, Jain, Mondal, and Chadha]{sahoo2024systematic}
Pranab Sahoo, Ayush~Kumar Singh, Sriparna Saha, Vinija Jain, Samrat Mondal, and Aman Chadha.
\newblock A systematic survey of prompt engineering in large language models: Techniques and applications.
\newblock \emph{arXiv preprint arXiv:2402.07927}, 2024.

\bibitem[Shankar et~al.(2024{\natexlab{a}})Shankar, Li, Asawa, Hulsebos, Lin, Zamfirescu-Pereira, Chase, Fu-Hinthorn, Parameswaran, and Wu]{shankar2024spade}
Shreya Shankar, Haotian Li, Parth Asawa, Madelon Hulsebos, Yiming Lin, J.~D. Zamfirescu-Pereira, Harrison Chase, Will Fu-Hinthorn, Aditya~G. Parameswaran, and Eugene Wu.
\newblock Spade: Synthesizing data quality assertions for large language model pipelines, 2024{\natexlab{a}}.

\bibitem[Shankar et~al.(2024{\natexlab{b}})Shankar, Zamfirescu-Pereira, Hartmann, Parameswaran, and Arawjo]{shankar2024validates}
Shreya Shankar, JD~Zamfirescu-Pereira, Bj{\"o}rn Hartmann, Aditya~G Parameswaran, and Ian Arawjo.
\newblock Who validates the validators? aligning llm-assisted evaluation of llm outputs with human preferences.
\newblock \emph{arXiv preprint arXiv:2404.12272}, 2024{\natexlab{b}}.

\bibitem[Skopek et~al.(2023)Skopek, Aralikatte, Gooding, and Carbune]{skopek-etal-2023-towards}
Ondrej Skopek, Rahul Aralikatte, Sian Gooding, and Victor Carbune.
\newblock Towards better evaluation of instruction-following: A case-study in summarization.
\newblock In Jing Jiang, David Reitter, and Shumin Deng, editors, \emph{Proceedings of the 27th Conference on Computational Natural Language Learning (CoNLL)}, pages 221--237, Singapore, December 2023. Association for Computational Linguistics.
\newblock \doi{10.18653/v1/2023.conll-1.16}.
\newblock URL \url{https://aclanthology.org/2023.conll-1.16}.

\bibitem[Srivastava et~al.(2022)Srivastava, Rastogi, Rao, Shoeb, Abid, Fisch, Brown, Santoro, Gupta, Garriga-Alonso, Kluska, Lewkowycz, Agarwal, Power, Ray, Warstadt, Kocurek, Safaya, Tazarv, Xiang, Parrish, Nie, Hussain, Askell, Dsouza, Slone, Rahane, Iyer, Andreassen, Madotto, Santilli, Stuhlmuller, Dai, La, Lampinen, Zou, Jiang, Chen, Vuong, Gupta, Gottardi, Norelli, Venkatesh, Gholamidavoodi, Tabassum, Menezes, Kirubarajan, Mullokandov, Sabharwal, Herrick, Efrat, Erdem, Karakacs, Roberts, Loe, Zoph, Bojanowski, Ozyurt, Hedayatnia, Neyshabur, Inden, Stein, Ekmekci, Lin, Howald, Orinion, Diao, Dour, Stinson, Argueta, Ram'irez, Singh, Rathkopf, Meng, Baral, Wu, Callison-Burch, Waites, Voigt, Manning, Potts, Ramirez, Rivera, Siro, Raffel, Ashcraft, Garbacea, Sileo, Garrette, Hendrycks, Kilman, Roth, Freeman, Khashabi, Levy, Gonz'alez, Perszyk, Hernandez, Chen, Ippolito, Gilboa, Dohan, Drakard, Jurgens, Datta, Ganguli, Emelin, Kleyko, Yuret, Chen, Tam, Hupkes, Misra, Buzan, Mollo, Yang, Lee, Schrader, Shutova,
  Cubuk, Segal, Hagerman, Barnes, Donoway, Pavlick, Rodol{\`a}, Lam, Chu, Tang, Erdem, Chang, Chi, Dyer, Jerzak, Kim, Manyasi, Zheltonozhskii, Xia, Siar, Mart'inez-Plumed, Happ'e, Chollet, Rong, Mishra, Winata, de~Melo, Kruszewski, Parascandolo, Mariani, Wang, Jaimovitch-L'opez, Betz, Gur-Ari, Galijasevic, Kim, Rashkin, Hajishirzi, Mehta, Bogar, Shevlin, Schutze, Yakura, Zhang, Wong, Ng, Noble, Jumelet, Geissinger, Kernion, Hilton, Lee, Fisac, Simon, Koppel, Zheng, Zou, Koco'n, Thompson, Wingfield, Kaplan, Radom, Sohl-Dickstein, Phang, Wei, Yosinski, Novikova, Bosscher, Marsh, Kim, Taal, Engel, Alabi, Xu, Song, Tang, Waweru, Burden, Miller, Balis, Batchelder, Berant, Frohberg, Rozen, Hern{\'a}ndez-Orallo, Boudeman, Guerr, Jones, Tenenbaum, Rule, Chua, Kanclerz, Livescu, Krauth, Gopalakrishnan, Ignatyeva, Markert, Dhole, Gimpel, Omondi, Mathewson, Chiafullo, Shkaruta, Shridhar, McDonell, Richardson, Reynolds, Gao, Zhang, Dugan, Qin, Contreras-Ochando, Morency, Moschella, Lam, Noble, Schmidt, He, Col'on, Metz,
  cSenel, Bosma, Sap, ter Hoeve, Farooqi, Faruqui, Mazeika, Baturan, Marelli, Maru, Quintana, Tolkiehn, Giulianelli, Lewis, Potthast, Leavitt, Hagen, Schubert, Baitemirova, Arnaud, McElrath, Yee, Cohen, Gu, Ivanitskiy, Starritt, Strube, Swkedrowski, Bevilacqua, Yasunaga, Kale, Cain, Xu, Suzgun, Walker, Tiwari, Bansal, Aminnaseri, Geva, Gheini, MukundVarma, Peng, Chi, Lee, Krakover, Cameron, Roberts, Doiron, Martinez, Nangia, Deckers, Muennighoff, Keskar, Iyer, Constant, Fiedel, Wen, Zhang, Agha, Elbaghdadi, Levy, Evans, Casares, Doshi, Fung, Liang, Vicol, Alipoormolabashi, Liao, Liang, Chang, Eckersley, Htut, Hwang, Milkowski, Patil, Pezeshkpour, Oli, Mei, Lyu, Chen, Banjade, Rudolph, Gabriel, Habacker, Risco, Milliere, Garg, Barnes, Saurous, Arakawa, Raymaekers, Frank, Sikand, Novak, Sitelew, Bras, Liu, Jacobs, Zhang, Salakhutdinov, Chi, Lee, Stovall, Teehan, Yang, Singh, Mohammad, Anand, Dillavou, Shleifer, Wiseman, Gruetter, Bowman, Schoenholz, Han, Kwatra, Rous, Ghazarian, Ghosh, Casey, Bischoff,
  Gehrmann, Schuster, Sadeghi, Hamdan, Zhou, Srivastava, Shi, Singh, Asaadi, Gu, Pachchigar, Toshniwal, Upadhyay, Debnath, Shakeri, Thormeyer, Melzi, Reddy, Makini, Lee, Torene, Hatwar, Dehaene, Divic, Ermon, Biderman, Lin, Prasad, Piantadosi, Shieber, Misherghi, Kiritchenko, Mishra, Linzen, Schuster, Li, Yu, Ali, Hashimoto, Wu, Desbordes, Rothschild, Phan, Wang, Nkinyili, Schick, Kornev, Tunduny, Gerstenberg, Chang, Neeraj, Khot, Shultz, Shaham, Misra, Demberg, Nyamai, Raunak, Ramasesh, Prabhu, Padmakumar, Srikumar, Fedus, Saunders, Zhang, Vossen, Ren, Tong, Zhao, Wu, Shen, Yaghoobzadeh, Lakretz, Song, Bahri, Choi, Yang, Hao, Chen, Belinkov, Hou, Hou, Bai, Seid, Zhao, Wang, Wang, Wang, and Wu]{Srivastava2022BeyondTI}
Aarohi Srivastava, Abhinav Rastogi, Abhishek Rao, Abu Awal~Md Shoeb, Abubakar Abid, Adam Fisch, Adam~R. Brown, Adam Santoro, Aditya Gupta, Adri{\`a} Garriga-Alonso, Agnieszka Kluska, Aitor Lewkowycz, Akshat Agarwal, Alethea Power, Alex Ray, Alex Warstadt, Alexander~W. Kocurek, Ali Safaya, Ali Tazarv, Alice Xiang, Alicia Parrish, Allen Nie, Aman Hussain, Amanda Askell, Amanda Dsouza, Ambrose Slone, Ameet~Annasaheb Rahane, Anantharaman~S. Iyer, Anders Andreassen, Andrea Madotto, Andrea Santilli, Andreas Stuhlmuller, Andrew~M. Dai, Andrew La, Andrew~Kyle Lampinen, Andy Zou, Angela Jiang, Angelica Chen, Anh Vuong, Animesh Gupta, Anna Gottardi, Antonio Norelli, Anu Venkatesh, Arash Gholamidavoodi, Arfa Tabassum, Arul Menezes, Arun Kirubarajan, Asher Mullokandov, Ashish Sabharwal, Austin Herrick, Avia Efrat, Aykut Erdem, Ayla Karakacs, B.~Ryan Roberts, Bao~Sheng Loe, Barret Zoph, Bartlomiej Bojanowski, Batuhan Ozyurt, Behnam Hedayatnia, Behnam Neyshabur, Benjamin Inden, Benno Stein, Berk Ekmekci, Bill~Yuchen Lin,
  Blake~Stephen Howald, Bryan Orinion, Cameron Diao, Cameron Dour, Catherine Stinson, Cedrick Argueta, C'esar~Ferri Ram'irez, Chandan Singh, Charles Rathkopf, Chenlin Meng, Chitta Baral, Chiyu Wu, Chris Callison-Burch, Chris Waites, Christian Voigt, Christopher~D. Manning, Christopher Potts, Cindy Ramirez, Clara Rivera, Clemencia Siro, Colin Raffel, Courtney Ashcraft, Cristina Garbacea, Damien Sileo, Daniel~H Garrette, Dan Hendrycks, Dan Kilman, Dan Roth, Daniel Freeman, Daniel Khashabi, Daniel Levy, Daniel~Mosegu'i Gonz'alez, Danielle~R. Perszyk, Danny Hernandez, Danqi Chen, Daphne Ippolito, Dar Gilboa, David Dohan, David Drakard, David Jurgens, Debajyoti Datta, Deep Ganguli, Denis Emelin, Denis Kleyko, Deniz Yuret, Derek Chen, Derek Tam, Dieuwke Hupkes, Diganta Misra, Dilyar Buzan, Dimitri~Coelho Mollo, Diyi Yang, Dong-Ho Lee, Dylan Schrader, Ekaterina Shutova, Ekin~Dogus Cubuk, Elad Segal, Eleanor Hagerman, Elizabeth Barnes, Elizabeth~P. Donoway, Ellie Pavlick, Emanuele Rodol{\`a}, Emma Lam, Eric Chu, Eric
  Tang, Erkut Erdem, Ernie Chang, Ethan~A. Chi, Ethan Dyer, Ethan Jerzak, Ethan Kim, Eunice~Engefu Manyasi, Evgenii Zheltonozhskii, Fanyue Xia, Fatemeh Siar, Fernando Mart'inez-Plumed, Francesca Happ'e, François Chollet, Frieda Rong, Gaurav Mishra, Genta~Indra Winata, Gerard de~Melo, Germ{\'a}n Kruszewski, Giambattista Parascandolo, Giorgio Mariani, Gloria~Xinyue Wang, Gonzalo Jaimovitch-L'opez, Gregor Betz, Guy Gur-Ari, Hana Galijasevic, Hannah Kim, Hannah Rashkin, Hannaneh Hajishirzi, Harsh Mehta, Hayden Bogar, Henry Shevlin, Hinrich Schutze, Hiromu Yakura, Hongming Zhang, Hugh~Mee Wong, Ian Ng, Isaac Noble, Jaap Jumelet, Jack Geissinger, John Kernion, Jacob Hilton, Jaehoon Lee, Jaime~Fern{\'a}ndez Fisac, James~B. Simon, James Koppel, James Zheng, James Zou, Jan Koco'n, Jana Thompson, Janelle Wingfield, Jared Kaplan, Jarema Radom, Jascha~Narain Sohl-Dickstein, Jason Phang, Jason Wei, Jason Yosinski, Jekaterina Novikova, Jelle Bosscher, Jennifer Marsh, Jeremy Kim, Jeroen Taal, Jesse Engel,
  Jesujoba~Oluwadara Alabi, Jiacheng Xu, Jiaming Song, Jillian Tang, Jane~W Waweru, John Burden, John Miller, John~U. Balis, Jonathan Batchelder, Jonathan Berant, Jorg Frohberg, Jos Rozen, Jos{\'e} Hern{\'a}ndez-Orallo, Joseph Boudeman, Joseph Guerr, Joseph Jones, Joshua Tenenbaum, Joshua~S. Rule, Joyce Chua, Kamil Kanclerz, Karen Livescu, Karl Krauth, Karthik Gopalakrishnan, Katerina Ignatyeva, Katja Markert, Kaustubh~D. Dhole, Kevin Gimpel, Kevin Omondi, Kory~Wallace Mathewson, Kristen Chiafullo, Ksenia Shkaruta, Kumar Shridhar, Kyle McDonell, Kyle Richardson, Laria Reynolds, Leo Gao, Li~Zhang, Liam Dugan, Lianhui Qin, Lidia Contreras-Ochando, Louis-Philippe Morency, Luca Moschella, Luca Lam, Lucy Noble, Ludwig Schmidt, Luheng He, Luis~Oliveros Col'on, Luke Metz, Lutfi~Kerem cSenel, Maarten Bosma, Maarten Sap, Maartje ter Hoeve, Maheen Farooqi, Manaal Faruqui, Mantas Mazeika, Marco Baturan, Marco Marelli, Marco Maru, Maria Jose~Ram'irez Quintana, Marie Tolkiehn, Mario Giulianelli, Martha Lewis, Martin
  Potthast, Matthew~L. Leavitt, Matthias Hagen, M'aty'as Schubert, Medina Baitemirova, Melody Arnaud, Melvin~Andrew McElrath, Michael Yee, Michael Cohen, Michael Gu, Michael~Igorevich Ivanitskiy, Michael Starritt, Michael Strube, Michal Swkedrowski, Michele Bevilacqua, Michihiro Yasunaga, Mihir Kale, Mike Cain, Mimee Xu, Mirac Suzgun, Mitch Walker, Monica Tiwari, Mohit Bansal, Moin Aminnaseri, Mor Geva, Mozhdeh Gheini, T~MukundVarma, Nanyun Peng, Nathan~A. Chi, Nayeon Lee, Neta Gur-Ari Krakover, Nicholas Cameron, Nicholas Roberts, Nick Doiron, Nicole Martinez, Nikita Nangia, Niklas Deckers, Niklas Muennighoff, Nitish~Shirish Keskar, Niveditha Iyer, Noah Constant, Noah Fiedel, Nuan Wen, Oliver Zhang, Omar Agha, Omar Elbaghdadi, Omer Levy, Owain Evans, Pablo Antonio~Moreno Casares, Parth Doshi, Pascale Fung, Paul~Pu Liang, Paul Vicol, Pegah Alipoormolabashi, Peiyuan Liao, Percy Liang, Peter Chang, Peter Eckersley, Phu~Mon Htut, Pi-Bei Hwang, P.~Milkowski, Piyush~S. Patil, Pouya Pezeshkpour, Priti Oli, Qiaozhu
  Mei, Qing Lyu, Qinlang Chen, Rabin Banjade, Rachel~Etta Rudolph, Raefer Gabriel, Rahel Habacker, Ramon Risco, Raphael Milliere, Rhythm Garg, Richard Barnes, Rif~A. Saurous, Riku Arakawa, Robbe Raymaekers, Robert Frank, Rohan Sikand, Roman Novak, Roman Sitelew, Ronan~Le Bras, Rosanne Liu, Rowan Jacobs, Rui Zhang, Ruslan Salakhutdinov, Ryan Chi, Ryan Lee, Ryan Stovall, Ryan Teehan, Rylan Yang, Sahib Singh, Saif~M. Mohammad, Sajant Anand, Sam Dillavou, Sam Shleifer, Sam Wiseman, Samuel Gruetter, Samuel~R. Bowman, Samuel~S. Schoenholz, Sanghyun Han, Sanjeev Kwatra, Sarah~A. Rous, Sarik Ghazarian, Sayan Ghosh, Sean Casey, Sebastian Bischoff, Sebastian Gehrmann, Sebastian Schuster, Sepideh Sadeghi, Shadi~S. Hamdan, Sharon Zhou, Shashank Srivastava, Sherry Shi, Shikhar Singh, Shima Asaadi, Shixiang~Shane Gu, Shubh Pachchigar, Shubham Toshniwal, Shyam Upadhyay, Shyamolima Debnath, Siamak Shakeri, Simon Thormeyer, Simone Melzi, Siva Reddy, Sneha~Priscilla Makini, Soo-Hwan Lee, Spencer Torene, Sriharsha Hatwar,
  Stanislas Dehaene, Stefan Divic, Stefano Ermon, Stella Biderman, Stephanie Lin, Stephen Prasad, Steven~T Piantadosi, Stuart~M. Shieber, Summer Misherghi, Svetlana Kiritchenko, Swaroop Mishra, Tal Linzen, Tal Schuster, Tao Li, Tao Yu, Tariq Ali, Tatsunori Hashimoto, Te-Lin Wu, Theo Desbordes, Theodore Rothschild, Thomas Phan, Tianle Wang, Tiberius Nkinyili, Timo Schick, Timofei Kornev, Titus Tunduny, Tobias Gerstenberg, Trenton Chang, Trishala Neeraj, Tushar Khot, Tyler Shultz, Uri Shaham, Vedant Misra, Vera Demberg, Victoria Nyamai, Vikas Raunak, Vinay~Venkatesh Ramasesh, Vinay~Uday Prabhu, Vishakh Padmakumar, Vivek Srikumar, William Fedus, William Saunders, William Zhang, Wout Vossen, Xiang Ren, Xiaoyu Tong, Xinran Zhao, Xinyi Wu, Xudong Shen, Yadollah Yaghoobzadeh, Yair Lakretz, Yangqiu Song, Yasaman Bahri, Yejin Choi, Yichi Yang, Yiding Hao, Yifu Chen, Yonatan Belinkov, Yu~Hou, Yu~Hou, Yuntao Bai, Zachary Seid, Zhuoye Zhao, Zijian Wang, Zijie~J. Wang, Zirui Wang, and Ziyi Wu.
\newblock Beyond the imitation game: Quantifying and extrapolating the capabilities of language models.
\newblock \emph{ArXiv}, abs/2206.04615, 2022.
\newblock URL \url{https://api.semanticscholar.org/CorpusID:263625818}.

\bibitem[Touvron et~al.(2023)Touvron, Lavril, Izacard, Martinet, Lachaux, Lacroix, Rozière, Goyal, Hambro, Azhar, Rodriguez, Joulin, Grave, and Lample]{touvron2023llama}
Hugo Touvron, Thibaut Lavril, Gautier Izacard, Xavier Martinet, Marie-Anne Lachaux, Timothée Lacroix, Baptiste Rozière, Naman Goyal, Eric Hambro, Faisal Azhar, Aurelien Rodriguez, Armand Joulin, Edouard Grave, and Guillaume Lample.
\newblock Llama: Open and efficient foundation language models, 2023.

\bibitem[Verga et~al.(2024)Verga, Hofstatter, Althammer, Su, Piktus, Arkhangorodsky, Xu, White, and Lewis]{verga2024replacing}
Pat Verga, Sebastian Hofstatter, Sophia Althammer, Yixuan Su, Aleksandra Piktus, Arkady Arkhangorodsky, Minjie Xu, Naomi White, and Patrick Lewis.
\newblock Replacing judges with juries: Evaluating llm generations with a panel of diverse models, 2024.

\bibitem[Wang et~al.(2023)Wang, Yu, Zeng, Yang, Wang, Chen, Jiang, Xie, Wang, Xie, Ye, Zhang, and Zhang]{Wang2023PandaLMAA}
Yidong Wang, Zhuohao Yu, Zhengran Zeng, Linyi Yang, Cunxiang Wang, Hao Chen, Chaoya Jiang, Rui Xie, Jindong Wang, Xingxu Xie, Wei Ye, Shi-Bo Zhang, and Yue Zhang.
\newblock Pandalm: An automatic evaluation benchmark for llm instruction tuning optimization.
\newblock \emph{ArXiv}, abs/2306.05087, 2023.
\newblock URL \url{https://api.semanticscholar.org/CorpusID:259108266}.

\bibitem[Wang et~al.(2022)Wang, Kordi, Mishra, Liu, Smith, Khashabi, and Hajishirzi]{Wang2022SelfInstructAL}
Yizhong Wang, Yeganeh Kordi, Swaroop Mishra, Alisa Liu, Noah~A. Smith, Daniel Khashabi, and Hannaneh Hajishirzi.
\newblock Self-instruct: Aligning language models with self-generated instructions.
\newblock In \emph{Annual Meeting of the Association for Computational Linguistics}, 2022.
\newblock URL \url{https://api.semanticscholar.org/CorpusID:254877310}.

\bibitem[Wei et~al.(2021)Wei, Bosma, Zhao, Guu, Yu, Lester, Du, Dai, and Le]{Wei2021FinetunedLM}
Jason Wei, Maarten Bosma, Vincent Zhao, Kelvin Guu, Adams~Wei Yu, Brian Lester, Nan Du, Andrew~M. Dai, and Quoc~V. Le.
\newblock Finetuned language models are zero-shot learners.
\newblock \emph{ArXiv}, abs/2109.01652, 2021.
\newblock URL \url{https://api.semanticscholar.org/CorpusID:237416585}.

\bibitem[Wei et~al.(2022)Wei, Wang, Schuurmans, Bosma, Xia, Chi, Le, Zhou, et~al.]{wei2022chain}
Jason Wei, Xuezhi Wang, Dale Schuurmans, Maarten Bosma, Fei Xia, Ed~Chi, Quoc~V Le, Denny Zhou, et~al.
\newblock Chain-of-thought prompting elicits reasoning in large language models.
\newblock \emph{Advances in neural information processing systems}, 35:\penalty0 24824--24837, 2022.

\bibitem[Zeng et~al.(2024)Zeng, Yu, Gao, Meng, Goyal, and Chen]{zeng2024llmbar}
Zhiyuan Zeng, Jiatong Yu, Tianyu Gao, Yu~Meng, Tanya Goyal, and Danqi Chen.
\newblock Evaluating large language models at evaluating instruction following.
\newblock In \emph{International Conference on Learning Representations (ICLR)}, 2024.

\bibitem[Zhang et~al.(2020)Zhang, Kishore, Wu, Weinberger, and Artzi]{zhang2020bertscore}
Tianyi Zhang, Varsha Kishore, Felix Wu, Kilian~Q. Weinberger, and Yoav Artzi.
\newblock Bertscore: Evaluating text generation with bert, 2020.

\bibitem[Zheng et~al.(2023{\natexlab{a}})Zheng, Chiang, Sheng, Li, Zhuang, Wu, Zhuang, Li, Lin, Xing, Gonzalez, Stoica, and Zhang]{Zheng2023LMSYSChat1MAL}
Lianmin Zheng, Wei-Lin Chiang, Ying Sheng, Tianle Li, Siyuan Zhuang, Zhanghao Wu, Yonghao Zhuang, Zhuohan Li, Zi~Lin, Eric~P. Xing, Joseph~E. Gonzalez, Ion Stoica, and Haotong Zhang.
\newblock Lmsys-chat-1m: A large-scale real-world llm conversation dataset.
\newblock \emph{ArXiv}, abs/2309.11998, 2023{\natexlab{a}}.
\newblock URL \url{https://api.semanticscholar.org/CorpusID:262084217}.

\bibitem[Zheng et~al.(2023{\natexlab{b}})Zheng, Chiang, Sheng, Zhuang, Wu, Zhuang, Lin, Li, Li, Xing, Zhang, Gonzalez, and Stoica]{zheng2023judging}
Lianmin Zheng, Wei-Lin Chiang, Ying Sheng, Siyuan Zhuang, Zhanghao Wu, Yonghao Zhuang, Zi~Lin, Zhuohan Li, Dacheng Li, Eric~P. Xing, Hao Zhang, Joseph~E. Gonzalez, and Ion Stoica.
\newblock Judging llm-as-a-judge with mt-bench and chatbot arena, 2023{\natexlab{b}}.

\bibitem[Zheng et~al.(2023{\natexlab{c}})Zheng, Chiang, Sheng, Zhuang, Wu, Zhuang, Lin, Li, Li, Xing, Zhang, Gonzalez, and Stoica]{Zheng2023JudgingLW}
Lianmin Zheng, Wei-Lin Chiang, Ying Sheng, Siyuan Zhuang, Zhanghao Wu, Yonghao Zhuang, Zi~Lin, Zhuohan Li, Dacheng Li, Eric~P. Xing, Haotong Zhang, Joseph Gonzalez, and Ion Stoica.
\newblock Judging llm-as-a-judge with mt-bench and chatbot arena.
\newblock \emph{ArXiv}, abs/2306.05685, 2023{\natexlab{c}}.
\newblock URL \url{https://api.semanticscholar.org/CorpusID:259129398}.

\bibitem[Zhou et~al.(2023)Zhou, Lu, Mishra, Brahma, Basu, Luan, Zhou, and Hou]{zhou2023instruction}
Jeffrey Zhou, Tianjian Lu, Swaroop Mishra, Siddhartha Brahma, Sujoy Basu, Yi~Luan, Denny Zhou, and Le~Hou.
\newblock Instruction-following evaluation for large language models.
\newblock \emph{arXiv preprint arXiv:2311.07911}, 2023.

\bibitem[Zhu et~al.(2023)Zhu, Wang, and Wang]{Zhu2023JudgeLMFL}
Lianghui Zhu, Xinggang Wang, and Xinlong Wang.
\newblock Judgelm: Fine-tuned large language models are scalable judges.
\newblock \emph{ArXiv}, abs/2310.17631, 2023.
\newblock URL \url{https://api.semanticscholar.org/CorpusID:264490588}.

\bibitem[Zhuang et~al.(2023)Zhuang, Yu, Wang, Sun, and Zhang]{Zhuang2023ToolQAAD}
Yuchen Zhuang, Yue Yu, Kuan Wang, Haotian Sun, and Chao Zhang.
\newblock Toolqa: A dataset for llm question answering with external tools.
\newblock \emph{ArXiv}, abs/2306.13304, 2023.
\newblock URL \url{https://api.semanticscholar.org/CorpusID:259243960}.

\end{thebibliography}

\appendix
\onecolumn
In this appendix, we include additional information on our work. This includes the prompts used in generating the dataset, distributions of domains and assertion criteria categories, example prompts in our dataset, a datasheet, and model cards.

\section{Prompt Template Analysis}
\label{app:prompthub}

In this section, we describe the prompts we used to identify the domain for each prompt template, the prompts used to generate the assertion criteria for each step of our workflow, and the distributions of our LLM-generated assertion criteria.

\subsection{Prompts and Analysis for Domain Categorization}
\label{app:prompthubdomain}

To categorize the prompt templates in \promptevals, we used the following prompts for querying GPT-4o. First, \texttt{DOMAIN\_CATEGORIZE\_1} was applied to each prompt template in parallel to generate fine-grained level 1 categories, resulting in 974 unique domains. Next, GPT-4o was queried once to consolidate these into 50 aggregate level 2 categories, which were sanity-checked for accuracy. Then, \texttt{DOMAIN\_CATEGORIZE\_2} was used to map each prompt template to one of the 50 predefined level 2 categories. We queried GPT-4o once more to come up with a set of level 3 categories (the highest level), and we manually assigned each level 2 category to a good level 3 category.

\begin{mypromptbox}[label={cat1}]{DOMAIN\_CATEGORIZE\_1}
Here is my prompt template for a specialized task:

\begin{verbatim}
```json
{prompt_template}
```
\end{verbatim}

What domain does this prompt template pertain to? Limit your response to a single word or phrase, and be as specific and fine-grained as possible. Avoid generic domains like `natural language processing` and `artificial intelligence.` Return your response as a JSON object in ```json ``` markers, with the key ``field'' and the value being the word or phrase.
\end{mypromptbox}

\begin{mypromptbox}[label={cat2}]{DOMAIN\_CATEGORIZE\_2}
Here is my prompt template for a specialized task:

\begin{verbatim}
```json
{prompt_template}
```
\end{verbatim}

The domain of this prompt template is: \begin{verbatim}{field}\end{verbatim}

Given the following higher level domains: General-Purpose Chatbots, Workflow and Task Automation, Question-Answering Systems, Education and Academic Assistance, Content Summarization and Extraction, Information Retrieval and Management, Programming and Software Development, Customer Support and Service, Data Analysis and Visualization, Content Creation and Writing, Project Management, Psychotherapy and Mental Health, Entertainment and Gaming, Healthcare and Medicine, Financial Services and Analysis, E-Commerce and Retail, Legal and Compliance, Marketing and Sales, Coaching and Personal Development, AI Evaluation and Optimization, Translation and Multilingual Services, Creative Arts and Media, Data Management and Databases, Text Analysis and Processing, Customer Experience and Feedback, Technology and IT Support, Evaluation and Quality Assurance, Real Estate and Property Management, Research and Information Synthesis, Insurance and Risk Management, Interactive Assistance and Support, Business Intelligence and Strategy, Human Resources and Recruitment, Knowledge and Information Synthesis, Task Execution and Management, Evaluation of AI Systems, Data Visualization and Reporting, Entertainment and Interactive Systems, Question Generation and Optimization, Automation and Orchestration, Translation and Language Services, Digital Marketing and SEO, Financial Services and Advising, Programming and Development Assistance, Creative and Content Writing, Healthcare and Medical Services, Customer Experience and Support, Business and Strategy Development, Evaluation and Quality Assurance, Human Resources and Recruitment

What higher level domain does this prompt template pertain to? You must pick one from the list above. Return only a JSON object in ```json ``` markers, with the key ``domain'' and the value being the word or phrase from the list above.
\end{mypromptbox}

\begin{figure}
    \centering
    \includegraphics[width=0.9\textwidth]{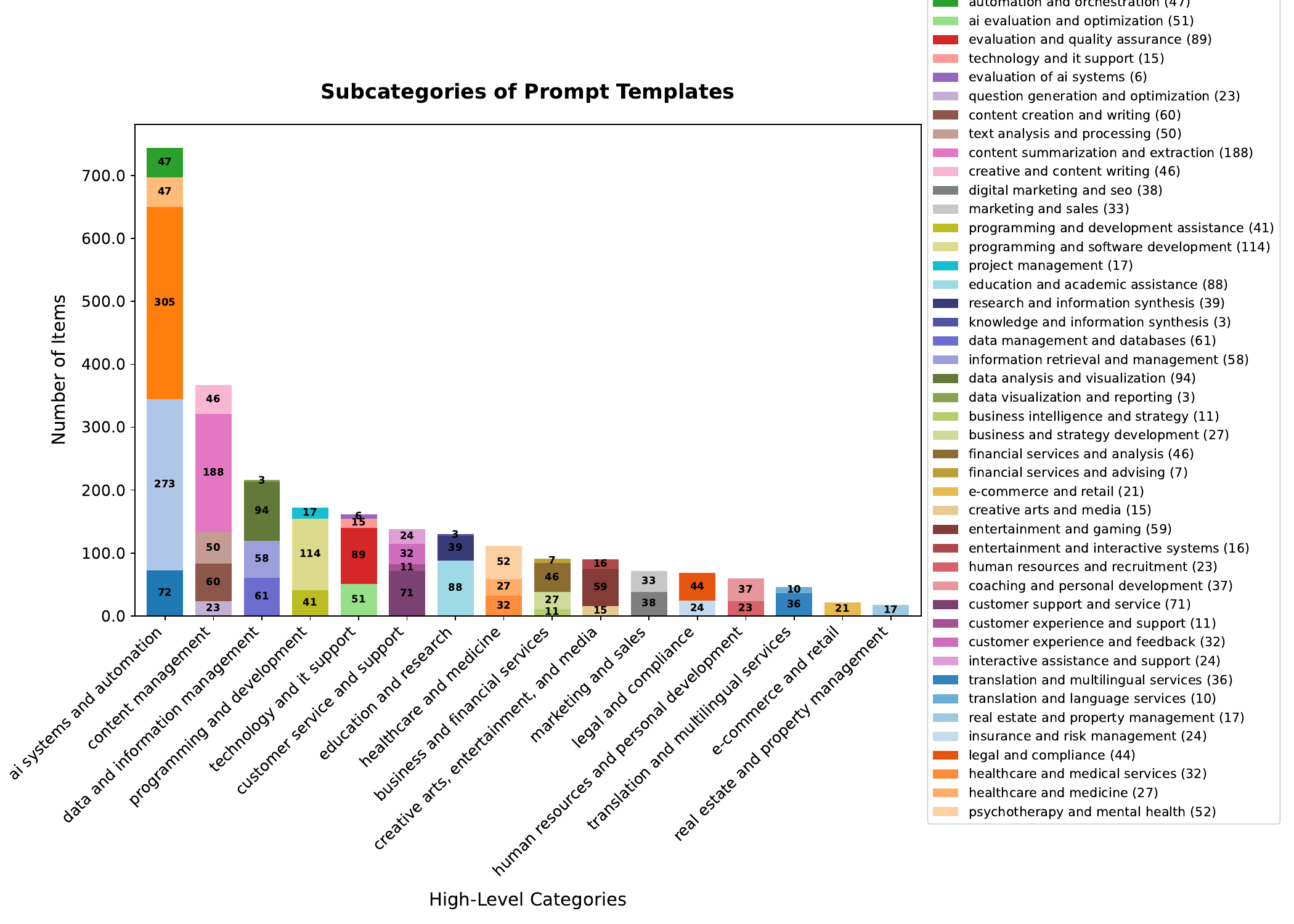}
    \caption{Distribution of Domains and Subdomains of Tasks Represented in \promptevals.}
    \label{fig:prompt-hub-domains}
\end{figure}

While our dataset in HuggingFace \url{https://huggingface.co/datasets/reyavir/PromptEvals} includes the finest-grained level 1 domains for each prompt template, \Cref{fig:prompt-hub-domains} shows the distribution of level 2 and level 3 domains across the dataset. Most prompt templates relate to AI systems and automation, as they are mainly prompt templates that generically evaluate or grade the quality of other LLM outputs, or they generically improve prompts to be more clear. The largest application of prompt templates is in content management (e.g., text summarization, creative writing), followed by data and information management (e.g., text to SQL).

\subsection{Prompts for LLM-Generated Criteria}

For each prompt template in the \promptevals, we use GPT-4o to generate a set of custom criteria that each LLM output should follow, or assertion criteria. Note that we use the term ``constraints'' here to match the terminology in Liu et al.~\cite{llmconstraints}. We refer the reader to their paper to read detailed descriptions of each constraint type.

Our prompt to generate the criteria is as follows:

\begin{mypromptbox}[label={constraintprompt}]{Prompt for Generating Custom Criteria} \label{subsec:ground-truth-criteria-prompt}
Here is my prompt template for a specialized task:

\begin{verbatim}
```json
{prompt_template}
```
\end{verbatim}

I want to write assertions for my LLM pipeline to run on all pipeline outputs. Here are some categories of constraints I may want the outputs to follow:

- **Structured Output**: Is there a requirement for the output to follow a standardized or custom format, such as markdown, HTML, or a JSON object?
- **Multiple Choice**: Does the output need to select from a predefined list of options?
- **Length Constraints**: Are there instructions regarding the targeted length of the output, such as the number of characters, words, or items in a list?
- **Semantic Constraints**:
  - **Excluding specific terms, items, or actions**: Are there terms, items, or actions that should be excluded from the output?
  - **Including or echoing specific terms or content**: Are there specific terms or content that should be included or echoed in the output?
  - **Covering or staying on a certain topic or domain**: Should the output cover or stay on a specific topic or domain?
  - **Following certain (code) grammar / dialect / context**: Are there requirements to follow certain (code) grammar, dialect, or context in the output?
- **Stylistic Constraints**: Are there requirements for the output to follow a certain style, tone, or persona?
- **Preventing Hallucination (Staying grounded and truthful)**: Should the output stay grounded and truthful, avoiding opinions, beliefs, or hallucinated outputs?
- **Preventing Hallucination (Adhering to Instructions without improvising unrequested actions)**: Should the output strictly adhere to any specific instructions provided, without including content that is not explicitly requested?

\{step\}\\
Return only your answer as a numbered list of strings.

\end{mypromptbox}

We updated the ``step'' input with a different prompt for each step of the workflow.\\
For step 1 (Generate initial criteria), the input was: \textit{Give me a list of constraints to implement for verifying the quality of my LLM output. Each item in the list should contain a string description of a constraint to check for and its corresponding type. Type names are: structured\_output, multiple\_choice, length\_constraints, exclude\_terms, include\_terms, stay\_on\_topic, follow\_grammar, stylistic\_constraints, stay\_truthful, adhere\_instructions.}

For step 2 (Add missing criteria), the input was: 

\textit{Here are some assertion constraints I want the outputs to follow: \{constraints\}\\
Add assertion constraints to the provided list. Add constraints that are stated in the prompt template and not already covered by an existing constraint. Return the combined list. Make sure the constraints are also followed by their corresponding categories.}

Where ``constraints'' was the output from the previous step.

For step 3 (Refine criteria), the input was: 

\textit{Here are some assertion constraints I want the outputs to follow: \{constraints\}\\
Remove any assertion constraints that are incorrect, redundant (or already covered by another constraint), not relevant to the prompt template, or difficult to validate. Make sure the constraints are also followed by their corresponding categories.}

Where ``constraints'' was the output from the previous step.

\subsection{Analysis of LLM-Generated Criteria}

\begin{figure*}[ht]
    \centering
    \includegraphics[width=0.9\textwidth]{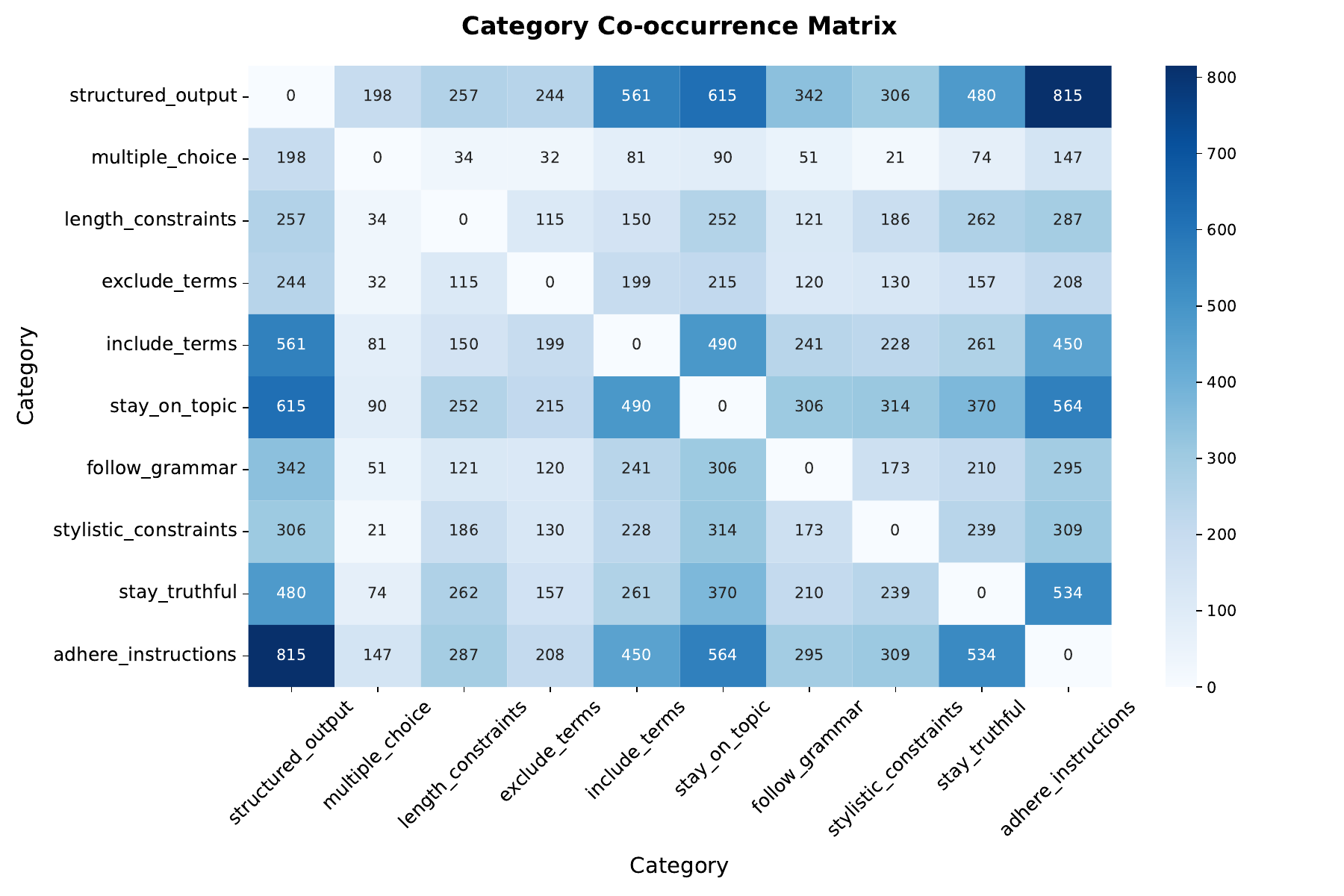}
    \caption{Constraint Type Co-Occurrence Matrix}
    \label{fig:constraints_cooccurrence}
\end{figure*}

\begin{figure}[ht]
    \centering
    \includegraphics[width=\textwidth]{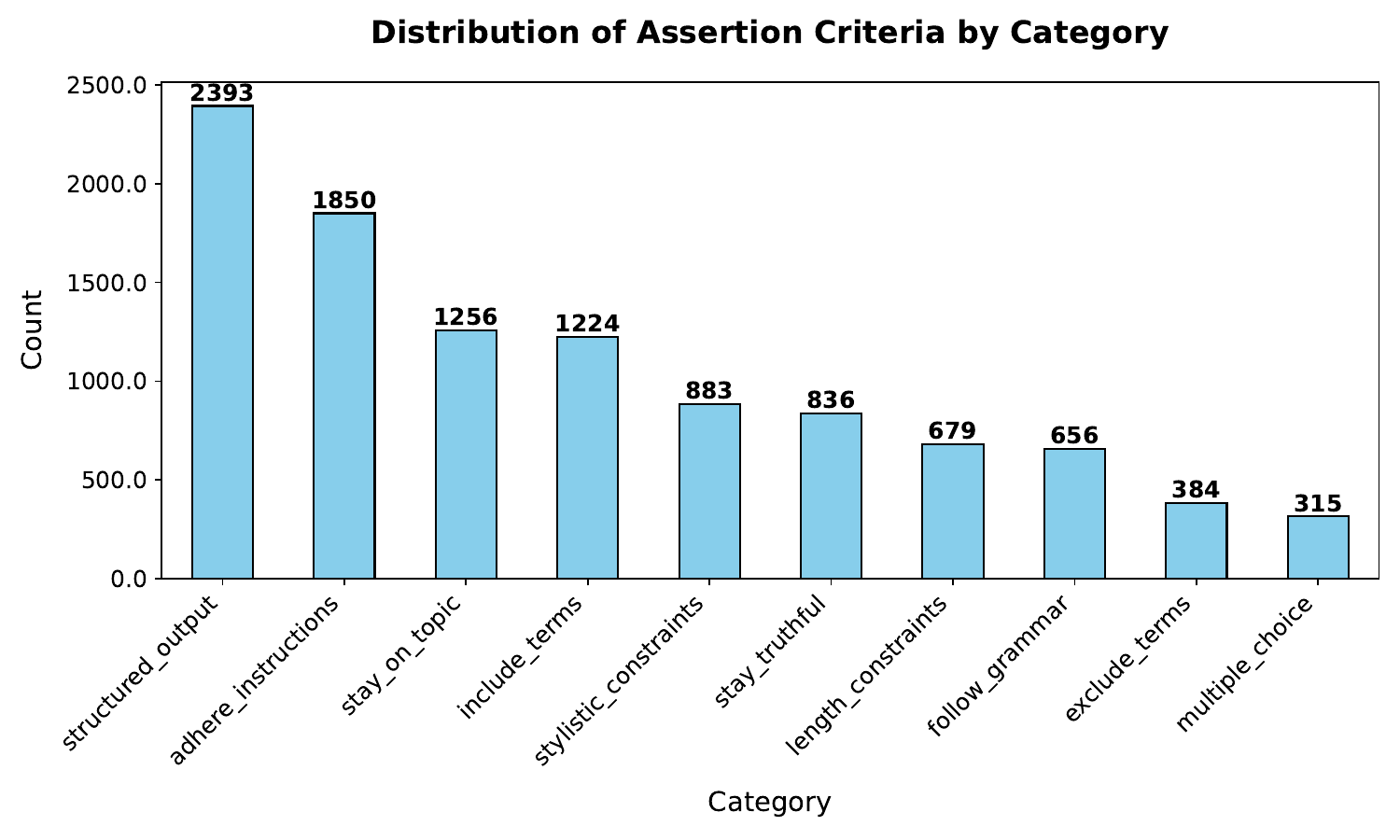}
    \caption{Distribution of Ground Truth Criteria by Type}
    \label{fig:constraints_by_category}
\end{figure}

We report distributions of the constraint types identified by GPT-4o in~\Cref{fig:constraints_by_category}, using the taxonomy from Liu et al.~\cite{llmconstraints}. \Cref{fig:constraints_by_category} shows the distribution of constraints across different categories, illustrating the prevalence of \texttt{structured\_output} type constraints. Additionally, Figure~\ref{fig:constraints_cooccurrence} shows a constraint type co-occurrence matrix. The top 5 co-occurring types are: \texttt{structured\_output} and \texttt{adhere\_instructions}, \texttt{structured\_output} and \texttt{stay\_on\_topic}, \texttt{adhere\_instructions} and \texttt{stay\_on\_topic}, \texttt{structured\_output} and \texttt{include\_terms}, and \texttt{stay\_truthful} and \texttt{adhere\_instructions}.

\section{Example Prompt Templates}


\subsection{Horse Racing Analytics} \label{subsec:horse-racing-prompt}
Prompt Template: ``SystemMessagePromptTemplate \\
ROLE:
You are a horse race analytic agent that explain a race detail with data and insight. You will receive user's question about a few horses' data, normally in numeric form. You have to first distinguish each horse's data, then answer user's question with the input and some professional's comments, your final output should be a decision of which horse is performing good or bad. \\
CONCEPTS:
You have to note these custom attributes before answering the question:
"Reborn" means that the horse has an insignificant drop in win odds, indicating that there are investment towards this horse when the win odds is not favourable. Showing that there are great increase in bettor's bet and confidence on this horse.
"SpeedPro" means the attribute given from a rating system on the horse's past running strategy. 
"Cold Door" means the comment rating on the horse from professional horse race commentor. \\
ACTION:
You have to analyze the separate horses and compare them with the data provided only, show which horse has the highest confidence with explanation. \\
HumanMessagePromptTemplate\\
I have a horse race with three horses participating, they has the record with : \{question\}. Now with the data supplied, summarize their potential performance.''

\section{Datasheet}
\begin{enumerate}
\item Why was the dataset created? (e.g., was there a specific intended task gap that needed to be filled?)\\
The dataset was created to be used in training or fine-tuning models in generating higher quality assertion criteria.
\item Who funded the creation of the dataset?\\
Lab sponsors.
\item What preprocessing/cleaning was done? (e.g., discretization or bucketing, tokenization, part-of-speech tagging, SIFT feature extraction, removal of instances) \\
The prompt template was extracted from the metadata and was added to the dataset. We removed any rows that resulted in 0 assertion criteria after the first step of our 3 step workflow.
\item If it relates to people, were they told what the dataset would be used for and did they consent?
If so, how? Were they provided with any mechanism to revoke their consent in the future or for certain uses? \\
Yes, the prompts are all from developers who consented to make their prompts public via a form. They can delete their prompts by submitting a delete request. We will semi-regularly update the Prompt Evals dataset to support the delete requests.

\item Will the dataset be updated? How often, by whom?\\
We plan to update the dataset yearly.
\end{enumerate}

\section{Model Cards}
\subsection{Fine-tuned Mistral}
\begin{enumerate}
\item Model Details. Basic information about the model. \\
– Person or organization developing model: MistralAI, and fine-tuned by the authors of this paper\\
– Model date: Base model released in September 2023, fine-tuned in July 2024 \\
– Model version: version 3\\
– Model type: decoder-only Transformer\\
– Information about training algorithms, parameters, fairness constraints or other applied approaches, and features: 7.3 billion parameters, fine-tuned by us using Axolotl (https://github.com/axolotl-ai-cloud/axolotl)\\
– Paper or other resource for more information:
https://arxiv.org/abs/2310.06825\\
– Citation details: https://openreview.net/forum?id=kW8wIpTgHF\\
– License: Apache 2.0 license\\
– Where to send questions or comments about the model: Reach out to the authors with any questions or comments\\
\item Intended Use. Use cases that were envisioned during development. (Primary intended uses, Primary intended users, Out-of-scope use cases)\\
Intended to be used by developers to generate high quality assertion criteria for LLM outputs, or to benchmark the ability of LLMs in generating these assertion criteria.
\item Factors. Factors could include demographic or phenotypic
groups, environmental conditions, technical attributes, or
others listed in Section 4.3.\\
We don’t collect any demographic, phenotypic, or others listed in Section 4.3, data in our dataset.
\item Metrics. Metrics should be chosen to reflect potential realworld impacts of the model. (Model performance measures, Decision thresholds, Variation approaches)\\
Metrics are defined in \Cref{sec:metrics}
\item Evaluation Data: Evaluated on \promptevals test set
\item Training Data: Fine-tuned on \promptevals train set
\item Quantitative Analyses (Unitary results, Intersectional results): See \Cref{tab:mistral_averages_per_domain}\\
\begin{table}[h!]
    \centering
    \footnotesize
    \begin{tabular}{lccccc}
        \toprule
        Domain & Similarity & Precision & Recall\\
        \midrule
        General-Purpose Chatbots & 0.8171 & 0.8023 & 0.8338 \\
        Question-Answering & 0.8216 & 0.8183 & 0.8255\\
        Text Summarization & 0.8785 & 0.8863 & 0.8725 \\
        Database Querying & 0.8312 & 0.8400 & 0.8234\\
        Education & 0.8599 & 0.8636 & 0.8564\\
        Content Creation & 0.8184 & 0.8176 & 0.8205\\
        Workflow Automation & 0.8304 & 0.8258 & 0.8351\\
        Horse Racing Analytics & 0.8216 & 0.8109 & 0.8336\\
        Data Analysis & 0.7865 & 0.7793 & 0.7952\\
        Prompt Engineering & 0.8534 & 0.8330 & 0.8755\\
        \bottomrule
    \end{tabular}
    \caption{Fine-Tuned Mistral Score Averages per Domain (for the 10 most represented domains in our test set)}
    \label{tab:mistral_averages_per_domain}
\end{table}

\item Ethical Considerations: See \Cref{sec:ethics}
\item Caveats and Recommendations: None
\end{enumerate}

\subsection{Fine-tuned Llama}
\begin{enumerate}
\item Model Details. Basic information about the model. \\
– Person or organization developing model: Meta, and fine-tuned by the authors of this paper \\
– Model date: Base model was released in April 18 2024, and fine-tuned in July 2024\\
– Model version: 3.1 \\
– Model type: decoder-only Transformer\\
– Information about training algorithms, parameters, fairness constraints or other applied approaches, and features: 8 billion parameters, fine-tuned by us using Axolotl (https://github.com/axolotl-ai-cloud/axolotl)\\
– Paper or other resource for more information:
https://arxiv.org/abs/2310.06825\\
– Citation details: https://openreview.net/forum?id=kW8wIpTgHF \\
– License: Meta Llama 3 Community License \\
– Where to send questions or comments about the model: Reach out to the authors with any questions or comments\\
\item Intended Use. Use cases that were envisioned during development. (Primary intended uses, Primary intended users, Out-of-scope use cases)\\
Intended to be used by developers to generate high quality assertion criteria for LLM outputs, or to benchmark the ability of LLMs in generating these assertion criteria.
\item Factors. Factors could include demographic or phenotypic
groups, environmental conditions, technical attributes, or
others listed in Section 4.3.\\
We don’t collect any demographic, phenotypic, or others listed in Section 4.3, data in our dataset.
\item Metrics. Metrics should be chosen to reflect potential realworld impacts of the model. (Model performance measures, Decision thresholds, Variation approaches)
Metrics are defined in \Cref{sec:metrics}
\item Evaluation Data: Evaluated on \promptevals test set
\item Training Data: Fine-tuned on \promptevals train set
\item Quantitative Analyses (Unitary results, Intersectional results): See \Cref{tab:llama_averages_per_domain}\\
\begin{table}[h!]
    \centering
    \footnotesize
    \begin{tabular}{lccccc}
        \toprule
        Domain & Similarity & Precision & Recall\\
        \midrule
        General-Purpose Chatbots & 0.8140 & 0.8070 & 0.8221\\
        Question-Answering & 0.8104 & 0.8018 & 0.8199\\
        Text Summarization & 0.8601 & 0.8733 & 0.8479\\
        Database Querying & 0.8362 & 0.8509 & 0.8228\\
        Education & 0.8388 & 0.8498 & 0.8282\\
        Content Creation & 0.8417 & 0.8480 & 0.8358\\
        Workflow Automation & 0.8389 & 0.8477 & 0.8304\\
        Horse Racing Analytics & 0.8249 & 0.8259 & 0.8245\\
        Data Analysis & 0.7881 & 0.7940 & 0.7851\\
        Prompt Engineering & 0.8441 & 0.8387 & 0.8496\\
        \bottomrule
    \end{tabular}
    \caption{Fine-Tuned Llama Score Averages per Domain (for the 10 most represented domains in our test set)}
    \label{tab:llama_averages_per_domain}
\end{table}
\item Ethical Considerations: See \Cref{sec:ethics}
\item Caveats and Recommendations: None
\end{enumerate}

\newpage

\end{document}